\documentclass{article} 
\usepackage{iclr2026_conference}

\usepackage{amsmath,amsfonts,bm}

\def\eqref#1{equation~\ref{#1}}

\def\1{\bm{1}}

\DeclareMathAlphabet{\mathsfit}{\encodingdefault}{\sfdefault}{m}{sl}
\SetMathAlphabet{\mathsfit}{bold}{\encodingdefault}{\sfdefault}{bx}{n}

\usepackage{xcolor}
\usepackage{array}
\usepackage{hyperref}
\usepackage{url}
\usepackage{booktabs}
\usepackage{graphicx}
\usepackage{amsmath}
\usepackage{newtxtext,newtxmath}
\usepackage{multirow}
\usepackage{subcaption}

\iclrfinalcopy

\newcommand{\gap}{\mathrm{gap}}
\newcommand{\dgap}{\widehat{\mathrm{gap}}}
\newcommand{\bgap}{\overline{\mathrm{gap}}}
\newcommand{\reenc}{\mathrm{reenc}}
\newcommand{\src}{\mathrm{src}}
\newcommand{\dem}{\mathrm{dem}}
\newcommand{\sigmastar}{\sigma^{\!*}}
\newcommand{\epsstar}{\varepsilon^{\!*}}

\definecolor{codeadd}{RGB}{0,90,180}
\newcommand{\addedline}[1]{\textcolor{codeadd}{#1}}

\title{When Does Training on Downscaled Images\\
Yield the Same Gradients?}

\author{Seunghyun Ji\\
Independent researcher\\
\texttt{standingbehindnv@gmail.com}
}

\begin{document}

\maketitle
\fancyhead{}
\renewcommand{\headrulewidth}{0.4pt}

\begin{abstract}
Diffusion transformers deliver strong image generation, but their
training cost grows superlinearly with resolution. Recent work
justifies training or sampling at reduced resolution on a spectral
premise: at high noise, a downscaled latent preserves almost the full
surviving signal. Whether a downscaled step also preserves the native
\emph{training gradient} signal, however, has remained unresolved. We
reduce how that signal changes under downscaling to two terms:
a noise-dependent term governed by the downscale
\emph{ratio}, which decays at high noise as the spectral premise
predicts, and a $\sigma$-independent floor governed by the target
grid's \emph{absolute token count}, carried by the compute graph
itself and removed by no noise level. The measured
$(\text{route},\sigma)$ map corroborates the account and uncovers
structure the spectral picture cannot express: on the $1024{\to}768$
route, a window ($0.65<\sigma<0.95$), predicted by no spectral
criterion at any tolerance, where the downscaled gradient stays
within a small margin of the native one. Training LoRA adapters with
downscaled steps restricted to the routes and noise windows the map
validates reduces training time by $14.6\%$ at a fixed step
budget while remaining near-native in weight space. Code is available at
\url{https://github.com/sorryhyun/anima_lora}.
\end{abstract}


\section{Introduction}
\label{sec:intro}

The cost of a forward pass in a diffusion transformer
\citep{peebles2023dit}, the backbone of current text-to-image
systems \citep{esser2024scaling,chen2024pixartalpha}, grows
superlinearly with resolution: token count grows quadratically with
edge length, and attention cost grows again in token count. This
cost structure makes the title question worth stating precisely: can
the \emph{gradient} of an adapter be computed on a downscaled latent
as a direct substitute for the native one, with the model, the
objective, and the $\sigma$ distribution unchanged and only
the spatial grid changed on some steps?

There is a well-studied reason to expect an affirmative answer
\emph{at high noise}. As the
noise level $\sigma$ grows, per-frequency signal-to-noise falls, and
high spatial frequencies are the first to be masked by the noise, so
that above a spectral crossover a downscaled latent carries
\emph{almost} the full surviving signal. Studies of this coupling, however, are
concentrated on the \emph{inference} side: scale-wise
distillation \citep{starodubcev2025swd}, spectral progressive
diffusion \citep{xiao2026spd}, and spectrally-guided noise schedules
\citep{esteves2026spectral} run high-noise \emph{sampling} steps at
reduced resolution and judge the premise by the generated output.
Where resolution is reduced during training
\citep{jin2024pyramidal,karras2018progressive,chen2024pixartalpha,hoogeboom2023simple},
the low-resolution stage is given its own, modified objective, and
success is again judged from the final samples. By the spectral
premise, substitution is safe \emph{whenever the noise has masked the
frequencies the coarse grid loses}.

This evidence, however, leaves the question open at \emph{training}
time: whether a downscaled step delivers the native gradient
direction under an unchanged objective. Output-level success observes
the substitution only through the confounds of a full training or
sampling pipeline, and a small difference in a quality score does not
imply a small difference in what a training step \emph{learns}. What
a training step consumes is the expected adapter gradient, and
substituting a downscaled, re-encoded latent for the native one at
training time perturbs more than the network's input: the regression
target carries the clean image at unit weight at every noise level,
and the compute graph itself changes with the grid. We therefore
decompose the gap between the downscaled and native gradients into
two terms, a noise-dependent term set by the downscale \emph{ratio}
and a $\sigma$-independent term set by the target grid's \emph{token
count}, and measure each directly.

The contributions of this paper are as follows:

\begin{enumerate}
\item \textbf{Two accounts of the gap, and a debiased measurement
that scores them} (\S\ref{sec:theory}, \S\ref{sec:evidence}). We
define the substitution gap at the gradient level and state two
accounts of it in the same units. The first is the gap
implied by the existing spectral argument, taken in its strongest
tolerance-parameterized form (SPD, \citealp{xiao2026spd}); it
depends on the downscale \emph{ratio} alone by construction. The
second, the \emph{gradient-perturbation} account, treats the
substitution as a perturbation of
both factors of $g = J^{\!\top}r$ and reduces, under mild,
realistic assumptions, to a route amplitude on a universal mismatch
curve \emph{plus} a $\sigma$-independent term set by the target
grid's token count. The two differ structurally before
any measurement (\S\ref{sec:spectral}), and the resulting
$(\text{route}, \sigma)$ map (\S\ref{sec:map}), read with each
route's finite-sample bias estimated and subtracted and against a
pre-specified margin, decides between them: it uncovers on the
$1024{\to}768$ route a window ($0.65 < \sigma < 0.95$) that the
spectral account does not predict at any tolerance. Scored head to
head on the same curves, the spectral account errs at $0.147$--$0.355$
RMSE, whereas ours predicts held-out routes at ${\sim}0.07$--$0.09$,
and the token-count law correctly predicts the $\sigma$-independent
term on both held-out routes.
\item \textbf{Selective low-resolution training} (\S\ref{sec:practical}).
Restricting the substitution to the window the map validates gives
an opt-in $\sigma$-conditional trainer that reduces the forward-compute
footprint by $15.1\%$ and training time by $14.6\%$ at a fixed step
budget, with a weight-space endpoint within kernel-noise reach of a
native retrain, rising to endpoint cosine $0.75$ when the
substitution is scheduled late at proportionally reduced saving
(Table~\ref{tab:yardstick}); a visual
counterpart (Fig.~\ref{fig:kaaiyuki}) confirms the endpoint renders
stay close to native.
\end{enumerate}

\section{Related work}
\label{sec:related}

\paragraph{Flow matching and DiT.}
Flow matching \citep{lipman2023flow} regresses a velocity field along
prescribed probability paths between noise and data; rectified flow
\citep{liu2023rectified} takes the paths to be straight lines, giving
the linear noising $z_\sigma = (1-\sigma)x + \sigma\epsilon$ and the
constant target $v = \epsilon - x$ used throughout, the training
objective of current large text-to-image systems
\citep{esser2024scaling}. The Diffusion Transformer
\citep{peebles2023dit} replaces the U-Net denoiser with a transformer
over patchified latent tokens, so a spatial grid enters the network
only as a token sequence: token count grows quadratically with edge
length and attention cost quadratically again in token count, the cost
structure that makes resolution the expensive axis. Spatial position is
now widely carried by a grid-dependent coordinate system, most commonly
rotary embeddings \citep{su2024roformer}.

\paragraph{Scale-wise and progressive-resolution diffusion.}
Progressive growing \citep{karras2018progressive} and resolution
curricula \citep{chen2024pixartalpha,hoogeboom2023simple} train at
increasing resolution as a schedule, each low-resolution stage carrying
its own stage-specific objective. A more recent family ties resolution
to the noise level itself, on the spectral premise above, and has so
far been studied on the \emph{inference} side. Scale-wise distillation
(SwD) \citep{starodubcev2025swd} states the claim directly (above the
crossover, the noised downscaled latent is close in distribution to the
downscaled noised latent) and samples early steps at reduced
resolution. Spectral Progressive Diffusion (SPD) \citep{xiao2026spd}
gives the argument its sharpest available form: a
tolerance-parameterized per-frequency activation time under a diagonal
Gaussian spectral model, a criterion on the per-band output of the
\emph{Bayes-optimal predictor} rather than merely on input SNR, from
which an inference-time resolution schedule follows. Spectrally-guided
noise schedules \citep{esteves2026spectral} apply the same reasoning
to schedule design; where the premise does enter training, as in
pyramidal flow matching \citep{jin2024pyramidal}, the objective is
restaged per pyramid level rather than left native.

\paragraph{Positional extension.}
Position interpolation \citep{chen2023pi}, banded/frequency-selective
variants (YaRN, \citealp{peng2024yarn}), and diffusion-specific dynamic
position extrapolation \citep{issachar2025dype,zhao2026sigma} rescale
rotary coordinates to evaluate a model at unseen grid sizes. We use
exact positional interpolation as a \emph{causal intervention} that
matches the downscaled grid's relative phase geometry to native, in
order to decompose the resolution-sensitivity floor, and adopt the
$\sigma$-gated band schedule of \citet{zhao2026sigma} as a
training-time refinement.
\section{Two accounts of what demotion costs}
\label{sec:theory}

This section fixes the estimand (\S\ref{sec:estimand}), then states
two accounts of it in the same units: a training-gap interpretation
built on the spectral account of the inference-side literature
(\S\ref{sec:spectral}), and the gradient-perturbation account
(\S\S\ref{sec:ouraccount}--\ref{sec:reduction}). The two
are not variants of one model: they disagree about which quantity
governs a route's cost, about what happens at pure noise, and about
whether a single parameter can order all routes.

\subsection{The estimand}
\label{sec:estimand}

\paragraph{Objective and noising.}
A flow-matching model regresses the constant-velocity transport between
a clean latent and pure noise. With clean latent $x$, noise
$\epsilon \sim \mathcal{N}(0, I)$, and noise level $\sigma \in (0,1]$,
the noised input is $z_\sigma = (1-\sigma)\,x + \sigma\,\epsilon$ and
the target is the velocity $v = \epsilon - x$; with caption condition
$c$ the loss is $\|\hat v_\theta(z_\sigma, \sigma, c) - v\|^2$.

\paragraph{Demotion.}
We write $e_0 {\to} e$ for a \emph{demotion}: the native latent at the
route's source edge $e_0$ is replaced with a coarser-grid one at its
target edge $e$.
With $y$ the image at its native resolution, $\mathcal{E}$ the VAE
encoder, and $\mathcal{R}_e$ the pixel-space downscale to the tier-$e$
bucket (native aspect ratio preserved),
$x_{\src} = \mathcal{E}(y) \longrightarrow
x_{\dem} = \mathcal{E}\big(\mathcal{R}_{e}(y)\big)$, yielding a latent
with proportionally fewer spatial tokens. The demoted step trains on
$z_\sigma = (1-\sigma)\,x_{\dem} + \sigma\,\epsilon$ with target
$v = \epsilon - x_{\dem}$, so the substitution changes both what the
network sees and what it is asked to predict, but only through the
spatial grid.

\paragraph{The estimand.}
For an arm $a \in \{\src, \reenc, \dem\}$, where the source arm
\emph{src} trains on the native-resolution latent $x_{\src}$, the demoted
arm \emph{dem} trains on $x_{\dem}$, and the control arm \emph{reenc} trains
on the native latent decoded and re-encoded at native resolution, let
$\bar g_a(\sigma) = \mathbb{E}_\epsilon[\nabla_\theta \mathcal{L}]$ be
the population per-bin mean adapter gradient of a query (image,
caption). The \emph{population demotion distance} of the route,
$d_{e_0 \to e}$, is
\begin{equation}
d_{e_0 \to e}(\sigma) \;=\; 1 - \cos\!\big(\bar g_{\src}(\sigma),\,
\bar g_{\dem}(\sigma)\big)
\;=\; \tfrac{1}{2}\,\big\|\hat g_{\src} - \hat g_{\dem}\big\|^2,
\label{eq:gapdef}
\end{equation}
with $\hat g$ the unit-normalized gradients. Route-level scalars carry
the \emph{route} as a subscript, abbreviated to the target edge alone
($d_e$) wherever the source is fixed by context; the
control's distance $d_{\reenc}$ carries its arm label instead, since it
changes no grid (full convention: Appendix~\ref{app:notation}). The
\emph{reported} quantity of every bin-resolved map and every fit in this
paper is the \emph{re-encoding excess}
\begin{equation}
\gap_e(\sigma) \;:=\; d_e(\sigma) \;-\; d_{\reenc}(\sigma),
\label{eq:excess}
\end{equation}
The control changes no grid, so its distance $d_{\reenc}$ prices the
VAE round trip alone; subtracting it isolates the part of $d_e$
attributable to the grid change. $\gap_e$ can be negative, $d_e$
cannot. Every
map in the body aggregates per example, taking the cosine per image
and then the mean; the batch-aggregate object that SGD follows in
batched training is a distinct estimand, treated in
Appendix~\ref{app:notation}.

\subsection{The spectral account}
\label{sec:spectral}

The spectral argument of \S\ref{sec:related} supplies the field's
working model of the same gap; we now restate it in the units of
Eq.~\ref{eq:gapdef}. Its strongest available form is the
tolerance-parameterized crossover of \citet{xiao2026spd}.

\paragraph{Demotion deletes a band.}
Write $u^{(\omega)}$ for the frequency-$\omega$ component of a field
$u$ in the latent's spatial Fourier decomposition, with $\omega$
measured in cycles per sample of the \emph{native} grid, so that the
native Nyquist frequency sits at $\tfrac12$. The downscale
$\mathcal{R}_e$ resamples the image with $e/e_0$ times fewer samples
per axis, and a grid at that density can represent no frequency above
half its own sampling rate; the resampler's anti-aliasing prefilter
removes, rather than aliases, whatever lies above. Setting aside the
VAE round trip, which the $\reenc$ control of \S\ref{sec:estimand} is
constructed to absorb, demotion is an ideal low-pass:
\begin{equation}
x_{\dem}^{(\omega)} \;=\;
\begin{cases}
x^{(\omega)}, & |\omega| < \omega_e,\\[3pt]
0, & |\omega| \ge \omega_e,
\end{cases}
\qquad
\omega_e \;=\; \frac{1}{2}\,\frac{e}{e_0}\,,
\label{eq:lowpass}
\end{equation}
mapping every band at or above the coarse grid's Nyquist frequency
$\omega_e$ to zero and acting as the identity on every band below it.

\paragraph{Carried to the training gradient.}
The noising $z_\sigma = (1-\sigma)\,x + \sigma\,\epsilon$ is linear
and hence diagonal in the spatial Fourier basis, so it acts on each
band independently. Modeling each clean-latent band as independently
Gaussian, $x^{(\omega)} \sim \mathcal{N}(0, P_\omega)$ with
$P_\omega$ the band's clean-latent power, \citet{xiao2026spd} attach
to each band an activation time: a crossover $t_\omega$ past which
the Bayes-optimal per-band velocity prediction agrees, to within a
tolerance $\delta$, with its data-independent limit
$\epsilon^{(\omega)}$.
The destroyed bands of Eq.~\ref{eq:lowpass} all sit at or above
$\omega_e$; on a decaying spectrum the lowest of them is the most
powerful, so it crosses last, and the safe-substitution boundary for
the whole route sits at $t_{\omega_e}$. The account attributes the
route's whole cost to those bands: whatever they contribute to the
residual, and through the backward pass to the adapter gradient,
perturbs $\bar g_{\dem}$ away from $\bar g_{\src}$. The implied
gradient gap is that contribution, gated at the Nyquist band's
activation time:
\begin{equation}
\gap_e^{\mathrm{spec}}(\sigma) \;=\;
S_e(\sigma)\,\mathbf{1}\!\big[\sigma < t_{\omega_e}\big],
\qquad
t_\omega \;=\;
\frac{1}{1 + \sqrt{\delta \,/\, \big(P_\omega\,(1 + P_\omega -
\delta)\big)}},
\label{eq:spd}
\end{equation}
where $S_e(\sigma)$ is the destroyed-band contribution to the distance
of Eq.~\ref{eq:gapdef}, and the tolerance $\delta$ is the account's
single free parameter, instantiated in Appendix~\ref{app:spectral}.
$S_e$ can be computed. At the residual level
the premise is complete: the Gaussian posterior mean is the classical
per-band Wiener shrinkage \citep{wiener1949}, so the cross-grid
mean-residual mismatch is confined to the destroyed band, ordered
across routes by destroyed-band energy, and follows in closed form
from the power spectrum (Appendix~\ref{app:posterior}). Carrying it
into the gradient units of Eq.~\ref{eq:gapdef} costs one calibrated
gain, since the map passes through the network's Jacobian, about which
a spectral model of the \emph{data} says nothing.
Appendix~\ref{app:spectral} instantiates this pipeline when
the account is scored.

Whatever the estimate of $S_e$, the account above commits to three
limits, fixed by the \emph{form} of Eq.~\ref{eq:spd} before any
gradient is measured.
\textbf{(a)~One parameter orders every route}: all boundaries move
together under the single tolerance, so the family is totally ordered.
\textbf{(b)~Ratio only}: the cut $\omega_e = \tfrac{1}{2}\,e/e_0$ of
Eq.~\ref{eq:lowpass} is a function of the downscale \emph{ratio} alone,
so the estimate is structurally blind to any absolute-size effect.
\textbf{(c)~No bias term}: for every route and every $\delta$
the predicted gap vanishes above a finite boundary, so no
$\sigma$-independent contribution is expressible.

\subsection{The gradient-perturbation account: what demotion perturbs}
\label{sec:ouraccount}

We start one level lower, from what the substitution touches
in the gradient itself. Differentiating the loss of
\S\ref{sec:estimand} gives, per draw and up to a constant,
\begin{equation*}
g \;=\; \nabla_\theta\, \tfrac12\big\|\hat v_\theta - v\big\|^2
\;=\; J^{\!\top} r,
\qquad
r \;=\; \hat v_\theta - v,
\qquad
J \;=\; \partial \hat v_\theta / \partial \theta,
\end{equation*}
with $r$ the prediction residual and $J$ the Jacobian through which
the compute graph turns a residual into an adapter gradient.
Demotion perturbs both factors: the residual, through the data the
objective consumes, and the Jacobian, through the compute graph. The
estimand is built from noise means, so split each factor into its
noise mean and fluctuation under the averaging that defines
$\bar g_a$ in \S\ref{sec:estimand}, $r_a = \bar r_a + \tilde r_a$ and
$J_a = \bar J_a + \tilde J_a$; the arm's mean gradient is then
\begin{equation*}
\bar g_a(\sigma)
\;=\; \mathbb{E}_\epsilon\!\big[J_a^{\!\top} r_a\big]
\;=\; \bar J_a^{\!\top}\,\bar r_a(\sigma)
\;+\; \mathbb{E}_\epsilon\!\big[\tilde J_a^{\!\top}\,\tilde r_a\big],
\qquad
\bar r_a(\sigma) \;=\; \mathbb{E}_\epsilon\big[\hat v_\theta - v\big],
\end{equation*}
and the \emph{mean residual} $\bar r_a$ is a velocity-shaped field
with a definite $\sigma$-curve for each arm.

Writing the demoted
arm's mean factors as source plus a perturbation,
$\bar r_{\dem} = \bar r + \Delta\bar r(\sigma)$ and
$\bar J_{\dem} = \bar J + \Delta J$ (bare bars denote the source arm),
subtracting the display across arms and expanding the product
(Appendix~\ref{app:branches}), the perturbation
$\delta g = \bar g_{\dem} - \bar g_{\src}$ decomposes as
\begin{equation}
\delta g \;=\;
\underbrace{\bar J^{\!\top}\Delta \bar r(\sigma)}_{\text{data branch }B_e}
\;+\; \underbrace{\Delta J^{\!\top}\, \bar r}_{\text{graph branch }C_e}
\;+\; \underbrace{\Delta J^{\!\top}\Delta \bar r + \cdots}_{\text{remainder }R_e},
\label{eq:branches}
\end{equation}
where the ellipsis collects the cross-arm difference of the
noise-covariance couplings
$\mathbb{E}_\epsilon[\tilde J_a^{\!\top}\tilde r_a]$.
$J_{\src}$ and $J_{\dem}$ act on different grids, so the branches are
not literal matrix products across arms; they are defined
\emph{operationally}, as the two independent perturbation directions in
the shared adapter-parameter space where both gradients live. Read
against Eq.~\ref{eq:branches}, the spectral account of
\S\ref{sec:spectral} keeps only the input-mediated part of the data
branch, gated at Nyquist, and sets the rest to zero.

Two interventions follow immediately from the decomposition, rather
than from experimental convention, and each isolates a branch. At
$\sigma = 1$ the input is pure noise in every arm, carrying no trace
of the image, so the input-mediated part of the data branch vanishes
by construction; the regression target, however, carries the clean
image at unit weight at every $\sigma$, so the mean residual retains
an image-dependent survivor,
$\bar r_a(1) = x_a - \mathbb{E}[x \mid c]$ up to the frozen model's
approximation error (Appendix~\ref{app:posterior}): the endpoint
reads the graph branch plus that target-mediated remnant of the data
branch. Zeroing the image in input \emph{and} target removes the
survivor as well: the arms then differ in nothing but the grid, every
data-mediated term vanishes by construction, and the x-zero probe
reads the graph branch alone. The one question the two probes cannot
settle, whether the survivor is \emph{resolvable} in the angular
estimand, is discharged in \S\ref{sec:ourscored}.

\subsection{The angular expansion and the two-term reduction}
\label{sec:reduction}

It remains to read Eq.~\ref{eq:branches} in the units of the estimand.
Split the perturbation $\delta g$ into a rescaling of $\bar g_{\src}$
and a rotation away from it, with $\kappa_\parallel$ and
$\kappa_\perp$ the two components in units of $\|\bar g_{\src}\|$.
The gap is then an exact saturating function of a single
rotation-to-scale ratio,
\begin{equation}
d_e \;=\; 1 - \frac{1}{\sqrt{1 + \kappa_{\mathrm{eff}}^2}}\,,
\qquad
\kappa_{\mathrm{eff}} \;=\; \frac{\kappa_\perp}{1 + \kappa_\parallel}\,,
\label{eq:exact}
\end{equation}
so only the orthogonal component moves the estimand, and the cosine
is \emph{blind} to a parallel rescaling. Expanding at small perturbation
and substituting Eq.~\ref{eq:branches} gives the \emph{four-term
angular expansion}: writing $u^{\perp}$ for the component of $u$
orthogonal to $\bar g_{\src}$,
\begin{equation}
d_e(\sigma) \;\approx\;
\underbrace{\frac{\big\|B_e^{\perp}\big\|^2}{2\,\|\bar g_{\src}\|^2}}_{\text{data share }S_e}
\;+\;
\underbrace{\frac{\big\|C_e^{\perp}\big\|^2}{2\,\|\bar g_{\src}\|^2}}_{\text{graph share }\Phi_e}
\;+\;
\underbrace{\frac{\big\langle B_e^{\perp},\, C_e^{\perp}\big\rangle}{\|\bar g_{\src}\|^2}}_{\text{projected interaction }I_e}
\;+\; R'_e,
\label{eq:fourterm}
\end{equation}
with both shares non-negative, an interaction that can carry either
sign, and a
remainder $R'_e$; the derivations and the remainder's expanded form
are discussed in Appendix~\ref{app:geometry}.

\paragraph{From the expansion to a measurable form.}
On the reported excess, Eq.~\ref{eq:fourterm} is exact but not yet an
instrument: no term in it is separately measurable per bin. The
excess itself does part of the work: the $\reenc$ arm changes no
grid, so $\Delta J_{\reenc} = 0$, its expansion degenerates to
$d_{\reenc} \approx S_{\reenc} + R'_{\reenc}$, and subtracting it
cancels only the re-encode component of the data share, passing the
graph share and interaction through untouched. The rest is carried by
four assumptions, each tested by a designated probe in
\S\ref{sec:evidence}:
\begin{itemize}
\item \textbf{(i)~Small remainder}: $R'_e$ negligible over the
claimed domain.
\item \textbf{(ii)~Negligible projected interaction}: $|I_e|$ small
against the retained shares.\footnote{Assumptions (i)--(ii) are not
unconstrained: both discarded terms obey Cauchy--Schwarz bounds
in the retained shares, the interaction by their geometric mean and
the remainder entering at half power in its own share, so (ii)
carries independent content only near share crossover
(Appendix~\ref{app:geometry}).}
\item \textbf{(iii)~Graph-relative stationarity}:
$\|C_e^{\perp}\| / \|\bar g_{\src}\|$ $\sigma$-constant, since
numerator and denominator share the residual's scale, so that the
graph share is flat in $\sigma$: a per-route constant, the
\emph{floor}.
\item \textbf{(iv)~Data-branch factorization}:
$\|B_e^{\perp}\| \approx a_e\,\|\Delta \bar r(\sigma)\|$, a
$\sigma$-independent route amplitude times the \emph{norm} of the
cross-grid mean-prediction-residual mismatch of Eq.~\ref{eq:branches},
asserted to be route-independent and directly measurable as such
(\S\ref{sec:ourscored}); the measured curve is an \emph{empirical}
closure of the mean residual's exact posterior form
(Appendix~\ref{app:posterior}), not an unconstrained fitted shape.
\end{itemize}
Under (i)--(iv) the expansion collapses to the \emph{two-term
reduction},
\begin{equation}
\gap_e(\sigma)
\;\approx\;
\underbrace{\tfrac{1}{2}\,a_e^2 \cdot \big(\|\Delta \bar r(\sigma)\| \,/\, \|\bar g_{\src}(\sigma)\|\big)^{2}}_{\text{data term}}
\;+\; \underbrace{\mathrm{RoPE}_e + \mathrm{Resid}_e}_{\text{graph term (the floor)}},
\label{eq:twoterm}
\end{equation}
a route amplitude $a_e$ on a measurable mismatch curve plus a
per-route constant; the supporting derivations are in
Appendix~\ref{app:geometry}. The floor is written with two parts
because a grid change touches two distinct things: the rotary
coordinate system, in which every band of the position embedding
accumulates phase at a different density across the image
($\mathrm{RoPE}_e$), and the non-positional graph statistics,
attention softmax over a different token count and the coarse graph's
capacity to approximate the fine graph's computation
($\mathrm{Resid}_e$).

The quadratic
power of the data term is fixed by the cosine geometry \emph{locally};
where the measured perturbation turns out large the licensed read is the
un-expanded parent form, Eq.~\ref{eq:exact} evaluated at the same
amplitude $a_e\,\|\Delta \bar r(\sigma)\|/\|\bar g_{\src}(\sigma)\|$
(recovering Eq.~\ref{eq:twoterm}'s data term in the small-mismatch
limit), whose data term
\emph{saturates} instead of overshooting.
The ledger is as follows: Eqs.~\ref{eq:exact} and~\ref{eq:fourterm}
are unconditional; Eq.~\ref{eq:twoterm} is conditional on the four
assumptions; and the coefficients $a_e$, $\|\Delta \bar r(\sigma)\|$,
$\|\bar g_{\src}(\sigma)\|$, and each route's floor are measured.

\paragraph{The estimand forces a debiased instrument.}
The exact angular link carries a metrological corollary. The estimand
is a cosine of population means; any measurement replaces $\bar g_a$
with a finite-draw average $g_a = \bar g_a + \xi_a$, and the draw noise
$\xi_a$, being uncorrelated with the mean, enters the geometry of
Eq.~\ref{eq:fourterm} as an orthogonal perturbation share of its own.
Every finite-draw cosine is therefore attenuated below its population
value, inflating the measured distance by a positive bias, and the
bias is arm-dependent: per-draw gradient variance grows as the token
count falls, so the inflation is larger on coarser grids and does not
cancel in the excess of Eq.~\ref{eq:excess}. The resulting artifact,
a positive offset persisting at every $\sigma$ and growing as the
target grid shrinks, matches the signature of the graph floor.
The same geometry dictates the remedy's form. The draw noise is
uncorrelated with the mean and independent across arms and draw sets,
so it cancels in every expected cross term while inflating every
squared norm: to first order the measured cosine factorizes,
$\cos(g_{\src}, g_a) \approx \lambda_{\src}\,\lambda_a
\cos(\bar g_{\src}, \bar g_a)$, one attenuation factor
$\lambda_a = \big(1 + \mathbb{E}\|\xi_a\|^2 /
\|\bar g_a\|^2\big)^{-1/2}$ per arm. Each factor is measurable in
place, since an arm's cosine with an independent redraw of itself,
$\cos_{\mathrm{self},a} = \cos(g_a, g_a')$, has expectation
$\lambda_a^2$; dividing by the geometric mean of the two self-cosines
is Spearman's classical correction for attenuation
\citep{spearman1904},
\begin{equation}
\hat c \;=\; \frac{\cos(g_{\src},\, g_{a})}
{\sqrt{\cos_{\mathrm{self},\src} \cdot \cos_{\mathrm{self},a}}},
\qquad
\dgap \;=\; 1 - \hat c,
\label{eq:debias}
\end{equation}
which cancels the finite-draw attenuation of numerator and
denominator to first order. The floor claim is therefore readable
only through the debiased gap $\dgap$;
\S\ref{sec:instrument} instruments Eq.~\ref{eq:debias} and validates
it.
\section{Experiments: both accounts, scored}
\label{sec:evidence}

This section scores both accounts of \S\ref{sec:theory} against one
measured object: debiased demotion-gap curves over the full $\sigma$
axis, on routes neither account saw. We build the instrument
(\S\ref{sec:instrument}), read the per-route map its curves support
(\S\ref{sec:map}), and then score both
accounts against the same curves: the spectral account at the
boundary and the curve level, the gradient-perturbation account term
by term (\S\ref{sec:scored}). Each subsection states the prediction
it discharges against criteria frozen before the runs.

\subsection{The instrument}
\label{sec:instrument}

\paragraph{Model and data.}
All measurements use Anima \citep{circlestone2025anima}, an open
DiT-based flow-matching text-to-image model (2B parameters, 28
transformer blocks, 3D rotary position embeddings, and the Qwen-Image
VAE, \citealp{wu2025qwenimage}), fine-tuned with LoRA adapters on an
illustration corpus. Images are bucketed by native aspect ratio into
resolution \emph{tiers} indexed by nominal edge length
$e \in \{512, 768, 896, 1024, 1280\}$; a tier-$e$ image occupies roughly
$(e/16)^2$ latent-patch tokens (${\sim}4{,}100$ at 1024, ${\sim}3{,}000$
at 896, ${\sim}2{,}160$ at 768, ${\sim}1{,}000$ at 512). The probe
adapter is a plain LoRA \citep{hu2022lora} checkpoint trained at
native tiers: low-rank adapters are the dominant fine-tuning vehicle
for models at this scale, and the trainer of \S\ref{sec:practical}
trains the same class.

\paragraph{The gradient probe.}
For each image and each of $B$ $\sigma$-bins we accumulate
adapter gradients over $D$ stratified noise draws per
arm and compare arms by cosine
similarity of the flattened accumulated gradients. The verdict run
uses $N = 40$ images and $D = 12$ draws per bin, on a segmented grid
of $14$ bins in $(0, 1)$, dense below $\sigma = 0.1$ and above $0.9$,
plus a $\sigma{=}1$ endpoint bin, with deterministic kernels. The
arms are: a
\textbf{floor}, two independent draw sets at native resolution, the
redraw null; the \textbf{reenc} control of \S\ref{sec:estimand}; one
\textbf{demote} arm per edge $e$; and, on debiased runs, per-arm
\textbf{self-floors}, a second independent draw set for every arm.
The $\sigma{=}1$ endpoint bin is a distinct probe \emph{mode}: its
input is pure $\epsilon$, so the input-mediated data term vanishes by
construction. An \textbf{x-zero} companion mode removes the image
from input \emph{and} target entirely. Both modes read only at the
endpoint, since with the $(1{-}\sigma)x$ term absent the lower bins
are off-manifold (Appendix~\ref{app:instrument}), and together they
carry the floor measurement of \S\ref{sec:scored}. The gap is in
cosine units, a scale-free mismatch \emph{fraction}, with bin-mean
SEM $0.01$--$0.07$ at $N{=}40$. Gap subtraction, not absolute
cosines, is the valid read: the floor itself moves with $\|g\|$
across bins. Every bin-mean curve must pass a split-half reliability
check over images, a discipline imposed after an earlier failure in
which per-image rankings from the same gradients had reliability
indistinguishable from zero.

\paragraph{Finite-draw attenuation and the debiased estimator.}
The arm-dependent attenuation of \S\ref{sec:reduction} is
substantial at our operating point — the natural estimate of
Eq.~\ref{eq:gapdef} also sets a redraw floor built from \emph{two}
finite-draw estimates against demote arms contributing one — and we
observed it directly: a first-order variance estimate predicts
spurious iso-direction endpoint gaps of
${\approx}\,0.02/0.05/0.15$ for $896/768/512$, and the uncorrected
$1024{\to}896$ endpoint gap indeed decays as
${+}0.100/{+}0.035/{-}0.016$ at $D = 4/8/16$, a clean $c/D$ decay
(Appendix~\ref{app:instrument}). Two corrections, used together,
remove it. First, \emph{self-floors}: every arm runs a second
independent draw set $g_a'$, restoring the floor's two-estimate
structure to every arm and supplying the per-arm self-cosines that
Eq.~\ref{eq:debias} consumes; the floor is precisely the source
arm's self-cosine, $\cos_{\mathrm{floor}} = \cos_{\mathrm{self},\src}$,
and the debiased gap $\dgap$ is computed per image and bin. Second,
\emph{draw-count extrapolation}: the attenuation is the $c/D$ decay
just measured, so fitting $\gap(D) = \gap_\infty + c/D$ over nested
draw subsets recovers the draw-limit gap directly. The two are
mutually checking, and the check passes: debiased fits come out
$D$-flat ($|c| \le 0.05$ on the $512$ route, against
$c \approx +0.29$ uncorrected), and the nested sweep extrapolates the
native redraw floor to a self-cosine of $1.005$ (bootstrap $95\%$ CI
$[0.994, 1.016]$). That is, the population per-bin native gradient is
a \emph{fixed direction}, and the entire redraw floor is finite-draw
noise; remaining validation detail is in
Appendix~\ref{app:instrument}.

\paragraph{Margins and resolution.}
A safety verdict needs two numbers, stated separately: a margin, the
largest excess considered practically equivalent and a property of
the question, and the instrument's resolution, which sets only the
power available to decide it. We fix the strict margin at
$\varepsilon = 0.02$, comparable to the re-encode control's own
confidence half-width (Table~\ref{tab:floor}); a verdict underpowered
there is stated at the smallest margin its bounds do clear. A route
is
\emph{safe at margin $\varepsilon$} in a bin where
the one-sided non-inferiority bound clears it,
\begin{equation}
\text{UB}_{95}\!\left(\bgap\right) \;<\; \varepsilon,
\label{eq:safe}
\end{equation}
and a \emph{window} claim over several bins requires the simultaneous
(Bonferroni-corrected) bounds to clear it in every bin. For a route
whose true excess is zero the paired estimate $\bgap$ fluctuates with
standard error $\mathrm{SE}(N, D, \sigma\text{-bin})$, so
$\epsstar = 1.645\,\mathrm{SE}$ is the \emph{verdict
resolution}: a cell with $\epsstar > \varepsilon$ is
\emph{underpowered} at the margin, and a crossing there is reported
but not counted as a safe verdict. At the verdict grid's operating point of $N{=}40$ and
$D{=}12$ per bin, the bin-level $\epsstar$ is $0.02$--$0.07$, so
window verdicts below are stated at the margins they do support
($0.09$ for the windows of \S\ref{sec:map});
endpoint-mode draw sweeps, with $D$ up to $64$, reach
$\epsstar \approx 0.01$--$0.02$, powered for the strict margin. All
verdict maps report the \emph{paired per-image excess}
$\gap_{e,i} = \dgap_{e,i} - \dgap_{\reenc,i}$, trimmed of non-finite
and $|\gap_{e,i}| > 1.5$ values, which cancels the shared per-image
draw structure and stays stable at mid-$\sigma$ where the unpaired
estimator is not; finite-draw cosines are compiler-kernel-path
sensitive, so all pairings are computed within one run. Every
verdict-carrying number in the main text is debiased; the
iso-severity, tier-transfer, and positional-interpolation probes of
\S\ref{sec:scored} remain raw-estimator reads, flagged where cited
(\S\ref{sec:limitations}).

\subsection{The measured $(\text{route}, \sigma)$ map}
\label{sec:verdict}
\label{sec:map}

\begin{figure}[t]
\centering
\includegraphics[width=\linewidth]{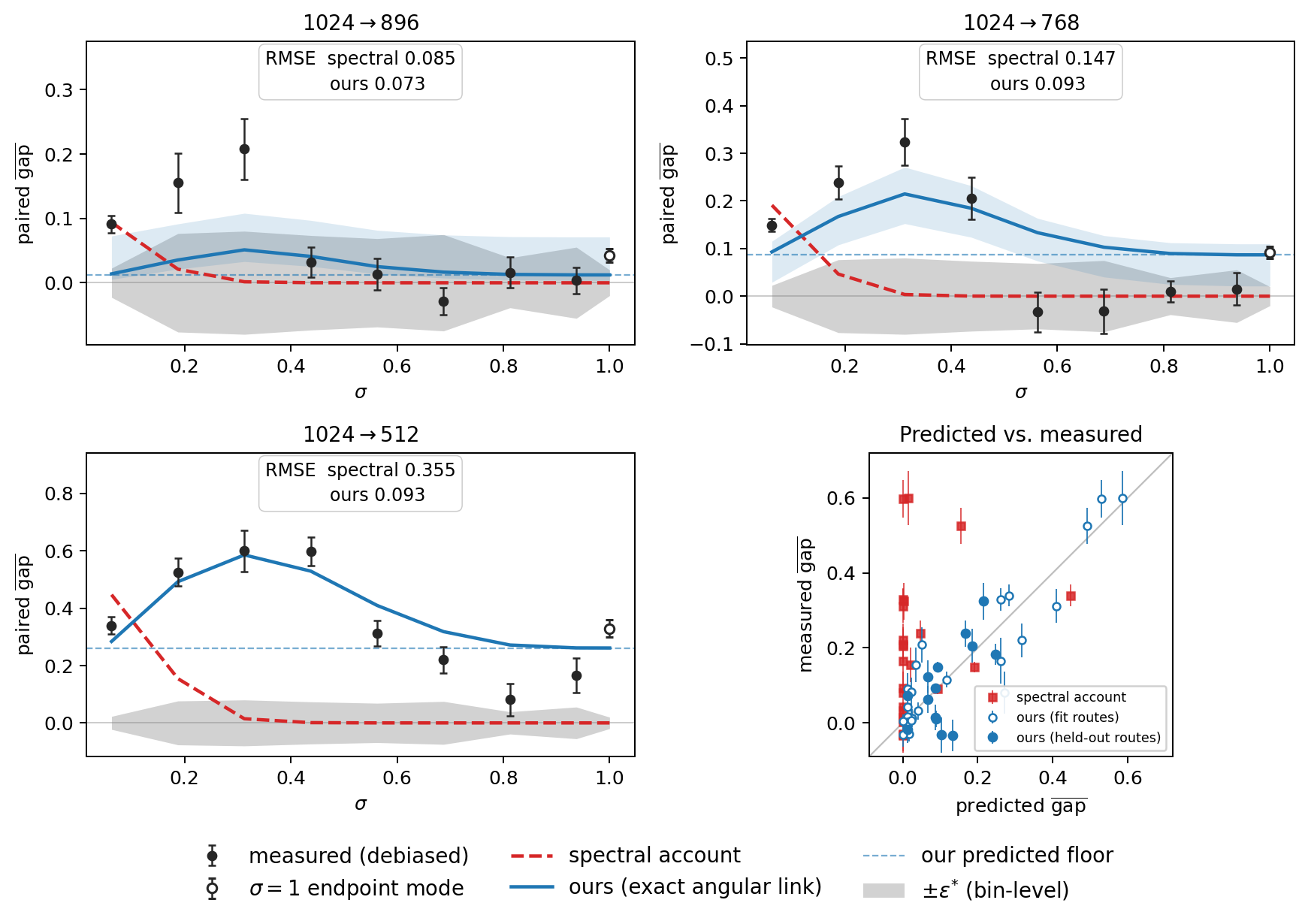}
\caption{\textbf{The measured map, and both accounts on the same
axes.} Per route, one set of measured debiased gap curves
(Eq.~\ref{eq:gapdef}, black; open markers detach the $\sigma{=}1$
endpoint bin, a distinct probe mode) carries the verdict of
\S\ref{sec:map}, read per bin against the gray band, the bin-level
$\pm\epsstar$ verdict resolution (\S\ref{sec:instrument}). The
same axes carry both predictions of \S\ref{sec:theory}: the spectral
account (red dashed), transported through the residual$\to$gradient
bridge at the measured anchor tolerance, the whole family scoring
alike (Appendix Fig.~\ref{fig:e83}); and the gradient-perturbation
account under the exact angular
link (blue, Eq.~\ref{eq:exact}; three-form comparison in Appendix
Fig.~\ref{fig:e51280}), with each route's predicted floor (blue
dashed; dashed-to-solid is the data term) and a $95\%$ full-pipeline
bootstrap band ($B{=}1000$; Appendix~\ref{app:instrument}).}
\label{fig:accounts}
\end{figure}

We measured the map directly: the measured (black) debiased gap curves
of Fig.~\ref{fig:accounts} carry the verdict, read per bin against the
gray $\pm\epsstar$ band per Eq.~\ref{eq:safe} (all three routes on one
axis, with the estimator context, in Appendix Fig.~\ref{fig:verdict};
per-bin numbers in Appendix Table~\ref{tab:debiasedmap});
\S\ref{sec:scored} scores both accounts on the same curves.
Throughout, \emph{safe} means the verdict of
Eq.~\ref{eq:safe} on the \emph{per-example} object: the sense in
which a demoted gradient is indistinguishable from a native one.
Route by route: $1024{\to}896$ (floor $0.02$--$0.04$) is
safe on the window $\sigma \in (0.5, 1)$ at margin $0.09$,
simultaneously over the seven window bins; above
$\sigma = 0.94$ the bins are underpowered at margins below
${\approx}\,0.05$ and the $\sigma{=}1$ endpoint carries a small real
gap (below), so the safe condition is $\sigma \in (0.5, 0.94]$
(\S\ref{sec:limitations}).
$1024{\to}768$ and $896{\to}768$ (floors $0.06$--$0.09$) are safe on no
window at the strict margin, but $1024{\to}768$ retains a narrow
low-excess window $\sigma \in (0.65, 0.95)$, with paired
excess $+0.02$--$0.04$ per bin, safe at margin $0.09$
simultaneously over its four window bins ($0.072$ pointwise). That
excess sits \emph{below} the route's own floor, a dip the two-term
reduction does not predict (the vector-resolved probe traces it to
negative data--graph interference; \S\ref{sec:limitations}); it is
the window \S\ref{sec:trainer}
stacks as a second route. Any route to $512$ (floor ${\approx}\,0.30$)
is \emph{unsafe} in all fifteen bins, the lower bound clearing the
margin in every one. Debiasing itself moved
verdicts in both directions: roughly half the $768$
route's high-$\sigma$ plateau was estimator bias, softening ``unsafe
at every noise level'' into the narrow window just stated, while the
$896$ endpoint shows a small \emph{real} gap that single-estimate
variance had buried ($+0.034 \pm 0.011$ in the verdict grid's
endpoint bin; $+0.019$ $[+0.010, +0.030]$ in the draw-limit sweep
that \S\ref{sec:scored} scores).

\subsection{Both accounts, scored}
\label{sec:scored}
\label{sec:specscored}
\label{sec:ourscored}

Both accounts are scored against one measured object: the debiased
gap curves of Fig.~\ref{fig:accounts}, read as \S\ref{sec:map} has
just set out. The spectral account is scored first, at the boundary
and then at the curve level; the gradient-perturbation account
follows, term by term.

\paragraph{The spectral account.}
The spectral account's prediction is monotone: the predicted gap is
concentrated at low $\sigma$ and vanishes at high $\sigma$, once
noise drowns the destroyed band (equal-power crossovers
$\sigma \approx 0.14/0.15/0.20$ for $896/768/512$, zero predicted gap
above). The measurement disagrees in both regimes: the transported
curves are near zero precisely where the measured curves peak (RMSE
$0.147$ at $768$, $0.355$ at $512$), and above the crossover the
floor persists (committed-region RMSE $0.027/0.049/0.218$). Neither
read depends on the tolerance or the residual-shape choice in the
transport (within $0.01$ RMSE; Table~\ref{tab:null},
Fig.~\ref{fig:e83}, Appendix~\ref{app:spectral}).

\paragraph{The data branch: a route-uniform residual
mismatch.}
Measuring the mean prediction residual
$\bar r = \mathbb{E}_\epsilon[\hat v - (\epsilon - x)]$ of
\S\ref{sec:reduction} per grid and
$\sigma$ shows a strong,
monotone $\sigma$-shape (cross-grid excess $0.89 \to 0.36$ in
relative $L_2$ from $\sigma{=}0.125$ to $1$) that is the \emph{same
for every route} within $\pm 0.02$, across routes whose gradient
floors span $0$ to $0.3$ (Appendix Table~\ref{tab:residual}). This
result refutes the residual-level prediction of \S\ref{sec:spectral}:
with no cross-frequency coupling, that mismatch is confined to the
destroyed band and ordered by its energy, which differs substantially
across these routes. Route-uniformity implies that the real mismatch
is carried by cross-band structure the diagonal model cannot express,
and that all route identity lives in $J$. The same contrast reads on
assumption~(iv): the measured route-uniformity is the shape
factorization that assumption asserts, established here directly
rather than assumed. The
decay is smooth, Wiener-like shrinkage with no hard gate at the
spectral crossover. How this monotone mismatch, read over the
U-shaped total gradient norm, yields the measured mid-$\sigma$-peaked
curves is a post-hoc consistency read deferred to
Appendix~\ref{app:instrument}.

\paragraph{The graph branch: the floor exists.}
At the $\sigma = 1$ \emph{endpoint} the input-mediated data term is
zero by construction (\S\ref{sec:ouraccount}); the debiased
draw-limit endpoint floors are
$+0.019 / +0.056 / +0.304$ for $896/768/512$ (Appendix
Table~\ref{tab:floor}),
and the $512$ value clears the pre-registered confirmation bar
($\dgap_\infty \ge 0.15$) on $12$ of $12$ probe images individually.
Three companion probes harden this number without adding to it. The
\emph{x-zero} probe reproduces the endpoint floors at every route
(Table~\ref{tab:floor}); a \emph{target-strength sweep} (target
$\epsilon - \alpha x$, $\alpha \in [0, 1]$) finds the paired debiased
excess $\alpha$-flat, discharging the survivor question of
\S\ref{sec:ouraccount} --- the target-mediated gradient is real but
demotion's change to it lands along $\hat g_{\src}$, to which the
angular estimand is blind (Eq.~\ref{eq:exact}); and a forward-only
probe of the model's caption-conditioned \emph{prior} dissociates the
prior from the floors' route ordering, placing the ordering in $J$.
The floor's split into $\mathrm{RoPE}_e + \mathrm{Resid}_e$
(Eq.~\ref{eq:twoterm}) is then read by one origin-side intervention:
evaluating the demoted grid at positionally-interpolated fractional
rotary coordinates, which matches its relative phase geometry to
native, \emph{erases} the $768$ endpoint floor and removes
${\sim}30\%$ of the $512$ floor, the in-band $896$ control untouched
(a raw-estimator probe; \S\ref{sec:limitations}). Re-anchored against
the debiased floors, $\mathrm{Resid}_{768} \approx 0$: what survives
the coordinate fix is the harsh route's
$\mathrm{Resid}_{512} \approx 0.2$ bulk, a genuine non-positional
graph residue, to which the capacity governor below attaches. (The
erasure itself is no training lever; the frequency-selective banded
alignment of \S\ref{sec:trainer} is the form that survives
in-window.) Constructions, anchors, the parallel-landing
decomposition behind the $\alpha$-flatness, and the floor ledger's
full numbers with their per-block depth localization are in
Appendix~\ref{app:instrument}.

\paragraph{The governors: ratio sets the amplitude, absolute size sets
the floor.}
With the floor isolated, what orders the routes? Two probes give a
clean division of labor, and together they decide the equal-ratio
disagreement, limit~(b) of \S\ref{sec:spectral}. \emph{Absolute
size, not ratio, sets the floor}: re-run one tier down, the route
$896{\to}768$ fails its pre-registered bar with a high-$\sigma$
residual ${\sim}2\times$ the $1024{\to}896$ plateau despite a
near-identical downscale ratio, while the two routes sharing the
${\sim}2{,}160$-token target grid, $1024{\to}768$ and $896{\to}768$,
land on the same floor despite different ratios (raw-estimator read,
the separation confirmed by the debiased same-grid floors; Appendix
Table~\ref{tab:phase1a}). Near-equal ratio with a different target
gives a different verdict; the same target with a different ratio
gives the same floor. This refutes any pure ratio rule for the floor,
and with it the spectral account's only degree of freedom. Moreover,
the floor grows monotonically as the \emph{target} grid shrinks
($0.02$--$0.04$ / $0.06$--$0.09$ / ${\approx}\,0.30$ at target
${\sim}3{,}000/2{,}160/1{,}000$ tokens), consistent with the reading
that the floor measures how well the coarse graph approximates the
fine graph's computation.
\emph{Ratio, not absolute size, sets the amplitude}: a synthetic
iso-severity route $1280{\to}1120$, with edge ratio exactly matched to
$1024{\to}896$ at $1.6\times$ its absolute target size, reproduces
the $1024{\to}896$ curve bin-for-bin, crossover included
(raw-estimator read; Appendix Table~\ref{tab:isoseverity}). Note what survives of
the spectral account here: ratio \emph{is} the right governor for the
data term; its error is not its governor but its scope, mistaking one
term for the whole gap.

\begin{table}[t]
\begin{minipage}[t]{0.615\linewidth}
\caption{Weight-space footprint:
$\cos(\Delta W_{\text{arm}}, \Delta W_{\text{native}})$ at matched
training seed, mean $\pm$ SD over three seed pairs, on two
single-artist corpora ($60$/$15$ training images). Training uses
deterministic kernels, including FlashAttention's deterministic
backward \citep{dao2022flashattention} (an identical retrain reads
$1.000$); the
non-det.\ row retrains native at the same seed \emph{without}
deterministic kernels (one twin pair per corpus), so it reads
run-to-run kernel nondeterminism alone. \emph{late}: demotion only in the last
$75\%$ of steps; the stacked row adds the $768$ low-excess window as
a second route ($1024{\to}768$ on $0.65{<}\sigma{<}0.95$, last $50\%$).
\textbf{Bold}: closest endpoints.}
\label{tab:yardstick}
\centering
\footnotesize
\setlength{\tabcolsep}{3pt}
\begin{tabular}{@{}lcc@{}}
\toprule
arm (vs.\ native, matched seed) & corpus A & corpus B \\
\midrule
native, same-seed non-det.\ retrain & $0.264$ & $0.427$ \\
\midrule
$896$, $\sigma{>}0.5$, aligned & $0.365 \pm .020$ & $0.432 \pm .026$ \\
\quad + late & $\mathbf{0.753 \pm .019}$ & $\mathbf{0.770 \pm .009}$ \\
\quad + late, stacked $768$ & $\mathbf{0.753 \pm .017}$ & $\mathbf{0.771 \pm .009}$ \\
$896$, every $\sigma$ & $0.183 \pm .009$ & $0.236 \pm .009$ \\
\bottomrule
\end{tabular}
\end{minipage}\hfill
\begin{minipage}[t]{0.355\linewidth}
\caption{The proposed step, in full. Lines 3--5 (colored) are the
entire change: the gate $\sigmastar{=}0.5$ and the route are read off
the measured map (\S\ref{sec:verdict}), line 5 is the optional
refinement, and lines 1--2 are only a reordering of a standard step,
such that $\sigma$ is drawn before the latent is fetched.}
\label{alg:sigmastep}
\centering
\footnotesize
\setlength{\tabcolsep}{2pt}
\begin{tabular}{@{}>{\ttfamily}r@{~}>{\ttfamily}l@{}}
\toprule
\multicolumn{2}{@{}l@{}}{\textbf{one LoRA training step}} \\
\midrule
1 & $\sigma \sim p_{\mathrm{train}}$ \\
2 & $x \leftarrow$ native latent of $y$ \\
\addedline{3} & \addedline{if $\sigma > \sigmastar$ and $y$ on-route:} \\
\addedline{4} & \addedline{\quad $x \leftarrow$ demoted latent of $y$} \\
\addedline{5} & \addedline{\quad rotary $\leftarrow$ banded$(\sigma)$} \\
6 & $\epsilon \sim \mathcal{N}(0, I)$ at shape$(x)$ \\
7 & $z \leftarrow (1{-}\sigma)x + \sigma\epsilon$ \\
8 & $v \leftarrow \epsilon - x$ \\
9 & step on $\|\hat v_\theta(z, \sigma, c) - v\|^2$ \\
\bottomrule
\end{tabular}
\end{minipage}
\end{table}

\section{$\sigma$-conditional training on the measured-safe route}
\label{sec:practical}
\label{sec:trainer}

As an applied example, we introduce the measured-safe corpus route
($1024{\to}896$) into a LoRA \citep{hu2022lora} trainer as an opt-in
path, read off the map as measured, with one stated liberty: the
gate is one-sided ($\sigma > \sigmastar$), so demoted steps also land
in the underpowered upper tail $\sigma \in (0.94, 1]$
(\S\ref{sec:limitations}). The intervention is
deliberately small: three lines inside an otherwise standard
flow-matching step, and no change to the objective, the optimizer, or
the noise density (pseudocode in Table~\ref{alg:sigmastep}). We
exercise it on a fixed-step grid: gated, late-scheduled (alone and
with the $768$ low-excess window stacked as a second route), and
gate-removed
arms $\times$ $3$ training seeds $\times$ two single-artist
illustration corpora of contrasting style, one flat and graphic, one
densely rendered (corpus A, $60$ training images; corpus B, $15$),
$480$ optimizer steps each, run with deterministic kernels (including
FlashAttention's deterministic backward,
\citealp{dao2022flashattention}) and paired RNG so
every arm sees identical draws. At fixed steps the always-gated arm
realizes a $-14.6\%$ training-time saving over native training on our
corpora; the late-scheduled arms demote only part of the run and save
proportionally less.

\textbf{$\sigma$-first draw.} The trainer draws each batch's $\sigma$
\emph{before} fetching latents, from the unchanged training density.
When every sample
in the batch is above the gate ($\sigma > 0.5$), the native
$1024$-tier latent is replaced with a cached demoted-grid counterpart
produced by the probe's own measured-safe recipe (pixel-space
downscale, VAE re-encode); everything downstream derives from the
substituted latent. Off-route images always train native, as does
validation.

\textbf{Banded rotary alignment, $\sigma$-gated.} Motivated by
\S\ref{sec:ourscored}, a frequency-selective alignment is applied on
demoted forwards only. It uses YaRN-style bands: a full
positional-interpolation stretch for rotary bands with few rotations
across the demoted extent, native spacing for high-frequency bands, a
ramp between, and thresholds scheduled by a sigmoid gate in $\sigma$
following \citet{zhao2026sigma}. Unlike uniform interpolation it is
\emph{not} off-manifold in-window: it passed both pre-registered
legs, erasing the static variant's low-$\sigma$ liability
($+0.064 \to +0.033 \pm 0.025$) while preserving its high-$\sigma$
gains (paired $-0.05$ vs plain demotion at $\sigma \ge 0.59$).

\begin{figure}[t]
\centering
\includegraphics[width=\linewidth]{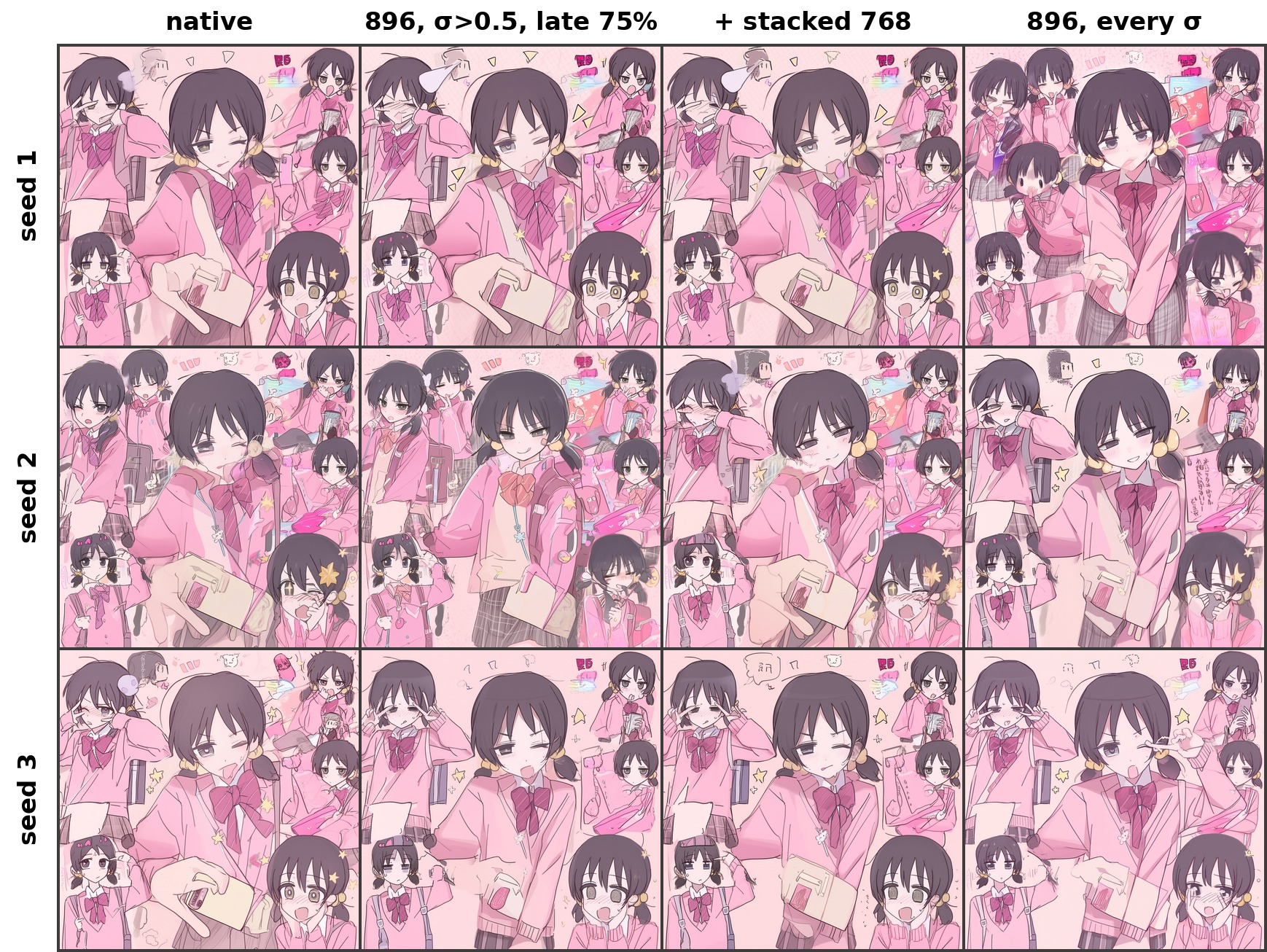}
\caption{Visual counterpart to Table~\ref{tab:yardstick}: matched
(prompt, noise-seed) renders from the arms' endpoints on a corpus-B
display prompt, $20$ steps; columns are four of the table's arms (the
unscheduled gated arm is omitted), rows the
three training seeds. Moving along a row (arms, seed fixed) changes
the render on the same visual order as moving down a column (training
seed, arm fixed); the gate-removed arm (rightmost) drifts the most,
matching its lowest cosine in Table~\ref{tab:yardstick}. The positive
prompt is given in Appendix~\ref{app:armsprompt}.}
\label{fig:kaaiyuki}
\end{figure}

\textbf{Weight-space endpoint footprint.} Table~\ref{tab:yardstick}
compares the trained adapters directly in weight space, as the cosine
between each arm's endpoint $\Delta W$ and its matched-seed native
twin's. Determinism makes the read exact: an identical retrain
reproduces its twin at $1.000$, and the same retrain with
non-deterministic kernels reads $0.264$/$0.427$ (corpus A/B),
run-to-run kernel nondeterminism alone. Against that reference, demoting
at \emph{every} $\sigma$ falls below the non-deterministic retrain
($0.183$/$0.236$), while demoting only above the gate
($\sigma{>}0.5$) matches or exceeds it ($0.365$/$0.432$).
Late scheduling, demotion only in the last $75\%$ of steps, brings
the endpoint very close to the native weights ($0.753$/$0.770$),
and, as the map's low-excess read of the $768$ window predicts,
stacking the $1024{\to}768$ route onto the late rule leaves the
cosine essentially unchanged ($0.753$/$0.771$). The visual
counterpart (Fig.~\ref{fig:kaaiyuki}) confirms the read: matched
(prompt, seed) renders from the late-scheduled arms are
nearly identical to native.

\section{Limitations}
\label{sec:limitations}

\textbf{One model family, one operating point.} All gradient probes ran
on one DiT (Anima) with one adapter checkpoint trained at native tiers.
A mixed-resolution-trained adapter might equalize (or widen) its own
gradients; probing such a checkpoint is the designated bound on how far
the map generalizes, as is re-probing $1280{\to}1024$ after fine-tuning
at the 1280 tier. The probe pool's relationship to the adapter's own
fine-tune set was measured rather than conceded: a controlled
$2{\times}2$ factorial (two LoRAs trained on opposite style clusters
under frozen membership manifests, probed on both) replicates the map's
shape on both checkpoints with the adapter$\times$probe-style
interaction below instrument resolution, while the absolute
redraw-floor level is checkpoint-dependent (Appendix~\ref{app:e7}).

\textbf{Gradient-level, not end-task.} The map is built from
accumulated per-step gradients; integration over an optimizer
trajectory is addressed by the exercise grid's weight-space read at
$480$ steps rather than production scale.

\textbf{Account resolution and domain.} The account predicts held-out
routes at ${\sim}0.07$--$0.09$ RMSE, not within $\epsstar$ (the
held-out protocol, its pre-registered gates, and the structural
failures it exposes are quantified in Appendix~\ref{sec:headtohead});
its
$A(\mathrm{ratio})$ governor rests on two distinct ratio values; the
reduction's domain excludes the $768$ mid-$\sigma$ window, whose
mechanism the vector-resolved probe has measured (negative
interference) but which the two-term reduction still does not predict; and
the floor law's functional form is unidentified at our operating
points. The graph-floor reading of the endpoint is
also estimand-specific: the target-mediated perturbation is real at the
vector level and lands parallel to the native gradient
(Appendix~\ref{app:instrument}), so a magnitude-sensitive estimand,
such as an un-normalized optimizer, could apportion the endpoint gap
differently than the angular read does.

\textbf{Instrument resolution and coverage.} ``Safe'' is one-sided
non-inferiority against a fixed margin ($\varepsilon = 0.02$ strict;
the map's windows are currently safe at $0.09$ simultaneous,
${\approx}\,0.07$ pointwise), with the
verdict resolution ($\epsstar = 0.02$--$0.07$ per bin;
${\approx}\,0.01$--$0.02$ at the endpoint sweeps) setting the power
available, not the margin. True gaps below $\epsstar$ are undetectable
by construction, $\sigmastar$ boundaries are localized only to a bin,
and every failure to reach a safe verdict on a window is $(N, D)$-relative: a
larger campaign could establish (or refute) the $768$ high-$\sigma$
window at the strict margin. The upper tail $\sigma \in (0.94, 1)$ of
the safe route is underpowered: its two non-endpoint bins resolve only
margins above ${\approx}\,0.05$, and the $\sigma{=}1$ endpoint
carries a small real gap ($+0.034 \pm 0.011$); an upper-tail sweep at
higher draw count is the designated probe to close the region, and
until it lands the safe condition is $\sigma \in (0.5, 0.94]$. Verdict stability under the per-image trim rule of
\S\ref{sec:instrument} has not been separately audited. The
debiasing campaign covered the corpus verdict grid; the $1280$-tier
iso-severity, $896$-tier transfer, depth-split, and PI-intervention
probes remain pre-debiasing reads, and the $1280$-tier curves enter the
held-out validation on that basis.

\section{Conclusion}
\label{sec:conclusion}

When does training on downscaled images yield the same gradient
direction? We answered by decomposing the demoted gradient's
perturbation into a ratio-governed data term and a
$\sigma$-independent, token-count-governed graph floor, and by
building the debiased estimator the question requires, since naive
finite-draw estimation manufactures the very floor signature under
test. Measured under that instrument, the spectral answer of the
inference-side literature (safe once noise masks the frequencies the
coarse grid loses) models only one part of one branch of what
demotion perturbs: the
network function itself is grid-calibrated, partly through its rotary
coordinate system and partly through an irreducible dependence of the
computation on token count. The gradient-perturbation account is predictive, not
merely descriptive, reaching held-out routes at ${\sim}0.07$--$0.09$
RMSE against the spectral family's $0.147$--$0.355$, and its map
licenses a selective trainer: on our model, $1024{\to}896$ at
$\sigma \in (0.5, 0.94]$ with a stackable $1024{\to}768$ window,
realizing $-14.6\%$ training time at fixed steps with a weight-space
endpoint within kernel-noise reach of a native retrain, rising to
cosine $0.75$ when demotion is late-scheduled at proportionally
reduced saving. We offer
the account and its instrument as the inexpensive, general test that any
future ``train part of the time at lower resolution'' proposal should
pass before it is believed.

\subsubsection*{Acknowledgements}
Portions of the text in this manuscript were drafted and edited with
the assistance of large language models (LLMs). The technical
contributions, experimental design, implementation, results, and
conclusions are those of the author, who has reviewed all generated
text and takes full intellectual responsibility for the content of
the paper.

\bibliography{references}
\bibliographystyle{iclr2026_conference}

\clearpage
\appendix

\section{Notation and derivations}
\label{app:derivations}

\subsection{Notation and estimand fine print}
\label{app:notation}

\paragraph{Arm labels versus route subscripts.}
One notational convention holds throughout. Arm labels ($\src$,
$\reenc$, $\dem$) mark \emph{whose} gradient, latent, or Jacobian an
object is; the subscript on every route-level scalar is the
\emph{route}, written in full ($d_{e_0 \to e}$) where several sources
appear, and abbreviated to the target edge alone wherever the source
edge is fixed by context, as it is in almost every measured
statement; the abbreviation is an edge value, never an arm label, so
it always instantiates numerically ($d_e$, $\Phi_{1120}$,
$\mathrm{Resid}_{768}$). The
control's distance $d_{\reenc}$ (Eq.~\ref{eq:gapdef} with
$\bar g_{\reenc}$ in place of $\bar g_{\dem}$) is the one exception that
carries an arm label: the control changes no grid and has no target
edge. Each table states which object it reports, $d_e$ or the excess
$\gap_e$.

\paragraph{The two aggregations.}
In a batched training scenario Eq.~\ref{eq:gapdef} admits two
aggregations over a probe set: per-example (cosine per image, then
mean; the object every map in the body reports) or batch-aggregate
(gradients summed over the batch before the cosine; the object SGD
actually follows). These are different estimands, since the
aggregation operator is part of the estimand, and no coefficient
fitted on one is guaranteed to transfer to the other. At second order the two
are related by a batch-size decomposition: writing each image's
demotion-induced gradient disagreement as a coherent mean $b$ plus a
zero-mean idiosyncratic deviation with covariance $\Sigma_\eta$, the
batch-aggregate distance at batch size $B$ is
\begin{equation*}
\mathbb{E}[d_B] \;\approx\;
\frac{\|P^{\perp} b\|^2}{2\|\mu\|^2}
\;+\;
\frac{\mathrm{tr}\!\big(P^{\perp} \Sigma_\eta P^{\perp}\big)}{2B\|\mu\|^2},
\end{equation*}
with $\mu$ the mean source gradient and $P^{\perp}$ the projector off
$\hat\mu$: an intercept (the coherent share, which never averages
out) plus a $1/B$ term. Monotone improvement with $B$ is not
automatic, since it would require an iid zero-mean disagreement model
we have not verified, but where we have measured the batch object it
is the more forgiving of the two: pooled-arm probes at pool size $4$
collapse the $1024{\to}896$ excess to ${\approx}\,0$ at every bin
$\sigma \ge 0.625$ and the $1024{\to}768$ excess at
$\sigma \ge 0.875$, the $1/B$ term dominating
there. What the collapse does not bound is the intercept, the
coherent drift no batch size removes. A debiased verdict grid with
pooled self-floors (pool $4$, $N{=}40$; paired arm$-$reenc gaps at
$B \in \{1, 4, 40\}$) bounds it: the $a + b/B$ fit's intercept is
${\leq}\,0.02$ ($1024{\to}896$) and ${\leq}\,0.03$ ($1024{\to}768$)
at $\sigma \geq 0.44$, but persists over the lower bins
($+0.04$--$0.09$, $+0.05$--$0.28$, and $+0.15$--$0.60$ for
$896/768/512$) — aggregation confers safety only where the per-example
map of \S\ref{sec:map} is already near-safe. The small real $896$
endpoint gap ($+0.042$ in this run's $B{=}1$ read) averages out in the
aggregate ($+0.001$).

\subsection{Derivation of the branch decomposition
(Eq.~\ref{eq:branches})}
\label{app:branches}

Fix a query (image, caption), a $\sigma$, and an arm $a$. Per draw
$\epsilon$, the adapter gradient of the $\tfrac12$-scaled loss is the
product $g_a(\epsilon) = J_a(\epsilon)^{\!\top} r_a(\epsilon)$ of
\S\ref{sec:ouraccount}. Split each factor into its noise mean and
fluctuation, $J_a = \bar J_a + \tilde J_a$ and
$r_a = \bar r_a + \tilde r_a$ with
$\mathbb{E}_\epsilon[\tilde J_a] = 0$ and
$\mathbb{E}_\epsilon[\tilde r_a] = 0$; the population mean gradient of
arm $a$ is then
\begin{equation}
\bar g_a \;=\; \mathbb{E}_\epsilon\!\big[J_a^{\!\top} r_a\big]
\;=\; \bar J_a^{\!\top} \bar r_a
\;+\; \mathbb{E}_\epsilon\!\big[\tilde J_a^{\!\top} \tilde r_a\big].
\label{eq:appmean}
\end{equation}
Write the demoted arm's mean factors as native plus a perturbation,
$\bar J_{\dem} = \bar J + \Delta J$ and $\bar r_{\dem} = \bar r + \Delta \bar r$
(bars without an arm subscript denote the native arm). Subtracting
Eq.~\ref{eq:appmean} across arms and expanding the product,
\begin{equation}
\delta g \;=\; \bar g_{\dem} - \bar g_{\src}
\;=\;
\underbrace{\bar J^{\!\top} \Delta\bar r}_{\text{data branch }B_e}
\;+\;
\underbrace{\Delta J^{\!\top}\, \bar r}_{\text{graph branch }C_e}
\;+\;
\underbrace{\Delta J^{\!\top} \Delta\bar r
\;+\;
\mathbb{E}_\epsilon\!\big[\tilde J_{\dem}^{\!\top}\tilde r_{\dem}\big]
- \mathbb{E}_\epsilon\!\big[\tilde J_{\src}^{\!\top}
\tilde r_{\src}\big]}_{\text{remainder }R_e}.
\label{eq:appbranches}
\end{equation}
This is an exact identity: no linearization is involved.
``First-order'' in the main text refers only to the content of $R_e$:
it collects every term that is a product of two perturbations
($\Delta J^{\!\top} \Delta\bar r$) or a difference of noise-covariance
terms (the last two). Eq.~\ref{eq:branches} is
Eq.~\ref{eq:appbranches} with the covariance difference abbreviated
into its ellipsis.

One remark delimits the identity's literal scope. When the two arms
share a grid (as for the re-encoding control), every operation above is
literal. Across grids, $J_{\src}$ and $J_{\dem}$ act on different
token counts, so the differences $\Delta J$ and $\Delta \bar r$ require
a fixed identification of the two output spaces (e.g.\ spectral
zero-padding of the coarse grid onto the fine grid's bands); any such
identification changes the split between $B_e$, $C_e$, and $R_e$ but
not their sum $\delta g$, which lives in the shared adapter-parameter
space regardless. This is why the main text defines the branches
\emph{operationally} (as the two independent perturbation directions
in adapter-parameter space, isolated by the designated probes of
\S\ref{sec:ourscored}) and treats Eq.~\ref{eq:branches} as fixing what
each branch collects rather than as a computable matrix formula.

\subsection{The mean residual as a posterior operator}
\label{app:posterior}

This appendix derives the posterior form of the fixed-image mean
residual, the exact split of the measured object into intrinsic
posterior bias plus model approximation error, and the Gaussian
closure referenced in \S\ref{sec:reduction}. Fix a caption $c$ and
grid $e$, write $a = 1 - \sigma$, and let $x \sim p_e(\cdot \mid c)$
be the grid-$e$ clean latent, $z_\sigma = a x + \sigma \epsilon$ with
$\epsilon \sim \mathcal{N}(0, I)$.

\paragraph{Bayes field and fixed-image residual.}
For the squared objective the population-optimal field is the
conditional mean of the target
$v = \epsilon - x = (z_\sigma - x)/\sigma$:
\begin{equation}
v_e^{*}(z, \sigma, c)
\;=\; \mathbb{E}[v \mid z_\sigma = z,\, c]
\;=\; \frac{z - m_e(z, c)}{\sigma},
\qquad
m_e(z, c) := \mathbb{E}[x \mid z_\sigma = z,\, c].
\label{eq:appbayesfield}
\end{equation}
Averaging over noise at fixed image
($\mathbb{E}_\epsilon[\epsilon] = 0$,
$\mathbb{E}_\epsilon[z_\sigma] = a x$) gives the posterior form
used in \S\ref{sec:reduction}:
\begin{equation}
\bar r_e^{*}(\sigma; x)
\;=\; \mathbb{E}_\epsilon\big[v_e^{*}(a x + \sigma \epsilon, \sigma, c)
      - (\epsilon - x)\big]
\;=\; \frac{x - \mathbb{E}_{z_\sigma \mid x}\big[m_e(z_\sigma,
      c)\big]}{\sigma}.
\label{eq:appposterior}
\end{equation}
With $p_{\sigma,e}(z \mid c)$ the noised marginal, Tweedie's identity
for the linear interpolant reads
$m_e(z, c) = \big(z + \sigma^2 \nabla_z \log p_{\sigma,e}(z \mid
c)\big)/a$ for $\sigma < 1$, so equivalently
\begin{equation}
\bar r_e^{*}(\sigma; x)
\;=\; -\frac{\sigma}{1 - \sigma}\,
\mathbb{E}_{z_\sigma \mid x}\big[\nabla_z \log
p_{\sigma,e}(z_\sigma \mid c)\big],
\label{eq:appscore}
\end{equation}
which makes the missing ingredient explicit: the numerical curve
requires the grid- and caption-conditional smoothed score, i.e.\ a
model of $p_e(x \mid c)$. At $\sigma = 1$, where $z = \epsilon$ is
independent of $x$, $m_e(z, c) \to \mathbb{E}[x \mid c]$ and
$\bar r_e^{*}(1; x) = x - \mathbb{E}[x \mid c]$. With $A_e$ the
alignment operator that places the source-grid residual on the demoted
grid (\S\ref{sec:instrument}), the Bayes-level cross-grid mismatch is
$\Delta \bar r_e^{*}(\sigma; x) = \bar
r_{\dem}^{*}(\sigma; x_{\dem}) - A_e \bar r_{\src}^{*}(\sigma;
x_{\src})$.

\paragraph{What the instrument measures.}
MSE optimality gives $\mathbb{E}[v^{*} - v \mid z, c] = 0$ and hence a
zero residual after a \emph{joint} average over $(x, \epsilon)$; it
does not zero the fixed-image average
$\mathbb{E}_\epsilon[v^{*} - v \mid x, c]$, which is
Eq.~\ref{eq:appposterior} and generally nonzero. Writing the trained
network as $\hat v_{\theta,e} = v_e^{*} + q_{\theta,e}$, the probe's
per-image object splits exactly as
\begin{equation}
\bar r_{\theta,e}(\sigma; x)
\;=\; \bar r_e^{*}(\sigma; x)
\;+\; \mathbb{E}_{z_\sigma \mid x}\big[q_{\theta,e}(z_\sigma, \sigma,
c)\big]:
\label{eq:appsplit}
\end{equation}
intrinsic posterior reconstruction bias, derivable in principle from
$p_e(x \mid c)$, plus genuinely model-specific approximation error.
The measured $\|\Delta \bar r(\sigma)\|$ therefore has definite
theoretical status, as an empirical closure of
Eq.~\ref{eq:appposterior}, but is not a pure model-error
difference, and no claim in the paper requires it to be one: the
two-term reduction consumes the measured curve as-is
(assumption~(iv)), and the trained-model residual is in any case the
exact $r$-factor that the LoRA gradient $g = J^{\!\top} r$ is built
from.

\paragraph{Gaussian closure.}
The simplest nontrivial closure is a full Gaussian model
$x \mid c \sim \mathcal{N}(\mu_e, C_e)$ per grid. With
$Q_e(\sigma) = a^2 C_e + \sigma^2 I$, the posterior mean is linear,
$m_e(z, c) = \mu_e + a C_e Q_e^{-1}(z - a \mu_e)$, and substituting
into Eq.~\ref{eq:appposterior} collapses (via
$I - a^2 C_e Q_e^{-1} = \sigma^2 Q_e^{-1}$) to
\begin{equation}
\bar r_e^{*}(\sigma; x) \;=\; \sigma\, Q_e(\sigma)^{-1} (x - \mu_e).
\label{eq:appgaussian}
\end{equation}
For paired source and demoted latents with covariance blocks
$C_{ss}, C_{dd}, C_{ds}$ and $L_s = \sigma Q_s^{-1}$,
$L_d = \sigma Q_d^{-1}$, the image-population mean squared mismatch is
\begin{equation}
\mathbb{E}\big\|\Delta \bar r_e^{*}\big\|^2
= \operatorname{tr}\!\big(L_d C_{dd} L_d^{\!\top}\big)
+ \operatorname{tr}\!\big(A_e L_s C_{ss} L_s^{\!\top}
  A_e^{\!\top}\big)
- 2 \operatorname{tr}\!\big(L_d C_{ds} L_s^{\!\top}
  A_e^{\!\top}\big),
\label{eq:apptrace}
\end{equation}
a $\sigma$-curve computable from paired \emph{clean latents only},
with the source--demote cross-covariance $C_{ds}$ load-bearing. A
\emph{diagonal Fourier} $C_e$ makes Eq.~\ref{eq:appgaussian} the
per-band Wiener shrinkage of \S\ref{sec:spectral}, confining the
mismatch to the destroyed band and ordering it by destroyed-band
energy: the residual-level prediction that route-uniformity
falsifies (\S\ref{sec:ourscored}). We tested whether any structured
second-order closure suffices, entirely on stored paired clean
latents (no denoiser forward): three nested covariance models
(diagonal spectral; $+$\,within-band channel blocks;
$+$\,octave-pair cross-band coupling), fitted on held-out-from-probe
images and scored on the probe set against the measured curves. All
three reproduce the curve's monotone $\sigma$-shape (Pearson
$0.94$--$0.97$), its route-uniformity, the near-zero re-encoding
control, and the $\sigma{=}1$ endpoint, but over-predict the
low-$\sigma$ amplitude by ${\sim}40\%$, failing a pre-registered
RMSE bar. Second-order data-only structure therefore recovers the
shape but not the level, and the essential missing ingredients are
the ones the identity names: caption-conditioning (the closure is
caption-marginal), non-Gaussian posterior structure, and the
$q_{\theta}$ term of Eq.~\ref{eq:appsplit}. The paper therefore
retains the measured $\|\Delta \bar r(\sigma)\|$ as its minimal
honest closure. A theory--instrument comparison must also reproduce
the instrument's estimand, the split-half-corrected \emph{relative}
$L_2$ distance of \S\ref{sec:instrument}, i.e.\
$\sqrt{\mathbb{E}\|\bar r_d - A_e \bar r_s\|^2}$ normalized by the
mean of the two arms' root-mean-square norms, rather than a raw
mismatch norm.

\subsection{The angular reduction: exact form and four-term expansion}
\label{app:geometry}

This appendix derives the two angular forms displayed in
\S\ref{sec:reduction}. Both follow from Eq.~\ref{eq:gapdef} and the
split of $\delta g$ by the native direction; no property of demotion
enters. Write $g = \bar g_{\src}$, $G = \|g\|$, $\hat g = g/G$, and
$u^{\perp} = u - (\hat g^{\!\top}u)\,\hat g$; decompose
$\delta g = G\,\kappa_\parallel\, \hat g + \delta g^{\perp}$ with
$\kappa_\parallel = \hat g^{\!\top}\delta g / G$ and
$\kappa_\perp = \|\delta g^{\perp}\| / G$.

\paragraph{The exact form.} The two components are orthogonal, so
$\|g + \delta g\|^2 = G^2 (1+\kappa_\parallel)^2 + G^2 \kappa_\perp^2$
and
\begin{equation}
\cos\big(g,\, g + \delta g\big)
\;=\; \frac{\hat g^{\!\top}(g + \delta g)}{\|g + \delta g\|}
\;=\; \frac{1 + \kappa_\parallel}
{\sqrt{(1+\kappa_\parallel)^2 + \kappa_\perp^2}}
\;=\; \frac{1}{\sqrt{1 + \kappa_{\mathrm{eff}}^2}}\,,
\qquad
\kappa_{\mathrm{eff}} = \frac{\kappa_\perp}{1 + \kappa_\parallel}\,,
\label{eq:appexact}
\end{equation}
valid whenever $1 + \kappa_\parallel > 0$ (the perturbation does not
reverse the gradient direction). The gap of Eq.~\ref{eq:gapdef} is
therefore exactly the saturating form displayed as Eq.~\ref{eq:exact}
in \S\ref{sec:reduction}. Since
$1 - \cos(\hat g_1, \hat g_2) = \tfrac12 \|\hat g_1 - \hat g_2\|^2$
for unit vectors, starting from either expression in
Eq.~\ref{eq:gapdef} lands on the same result.

\paragraph{The quadratic limit.} For small perturbations,
$1 - (1 + \kappa_{\mathrm{eff}}^2)^{-1/2}
= \tfrac12 \kappa_{\mathrm{eff}}^2 + O(\kappa_{\mathrm{eff}}^4)$ and
$\kappa_{\mathrm{eff}}^2 = \kappa_\perp^2\big(1 - 2\kappa_\parallel
+ O(\kappa_\parallel^2)\big)$, so
\begin{equation}
d_e \;=\; \frac{\|\delta g^{\perp}\|^2}{2 G^2}
\;+\; \underbrace{O(\kappa_\perp^2 \kappa_\parallel)
+ O(\kappa_\perp^4)}_{\text{beyond-quadratic}}.
\label{eq:appquad}
\end{equation}

\paragraph{The four-term expansion.} Substituting
$\delta g = B_e + C_e + R_e$ (Eq.~\ref{eq:appbranches}) into the leading
term and expanding the square,
\begin{equation*}
\|\delta g^{\perp}\|^2
= \|B_e^{\perp}\|^2 + \|C_e^{\perp}\|^2
+ 2\,\langle B_e^{\perp},\, C_e^{\perp}\rangle
+ \Big(2\,\langle (B_e{+}C_e)^{\perp},\, R_e^{\perp}\rangle
+ \|R_e^{\perp}\|^2\Big),
\end{equation*}
whose first three terms, divided by $2G^2$, are the data share $S_e$,
the graph share $\Phi_e$, and the projected interaction $I_e$ of the
four-term expansion, Eq.~\ref{eq:fourterm}. The remainder
$R'_e$ has an exact expansion: the
$R_e$-bearing terms above, the beyond-quadratic terms of
Eq.~\ref{eq:appquad}, and the parallel-component correction, each
suppressed by an extra power of perturbation size relative to the
retained terms.

\paragraph{From the four-term form to the excess.} The reported object
is the excess over the control. The $\reenc$ arm changes no grid, so
$\Delta J_{\reenc} = 0$, and both control terms built from it vanish
identically: its graph share $\Phi_{\reenc}$ and its projected
interaction $I_{\reenc}$ each carry a factor
$(\Delta J_{\reenc}^{\!\top}\bar r)^{\perp} = 0$. Writing
Eq.~\ref{eq:fourterm} for each arm and subtracting term by term,
\begin{align*}
\gap_e(\sigma)
\;&=\; d_e(\sigma) \;-\; d_{\reenc}(\sigma)\\
&\approx\;
\big(S_e - S_{\reenc}\big) \;+\; \big(\Phi_e - \Phi_{\reenc}\big)
\;+\; \big(I_e - I_{\reenc}\big) \;+\; \big(R'_e - R'_{\reenc}\big)
&&\text{Eq.~\ref{eq:fourterm}, each arm}\\
&=\;
\big(S_e - S_{\reenc}\big) \;+\; \Phi_e \;+\; I_e
\;+\; \big(R'_e - R'_{\reenc}\big),
&&\Phi_{\reenc} = I_{\reenc} = 0
\end{align*}
so the control collapses to
$d_{\reenc} \approx S_{\reenc} + R'_{\reenc}$, with $S_{\reenc}$ the
re-encode share (measured ${\approx}\,0$ where probed: the control row
of Table~\ref{tab:floor}). The subtraction cancels only the re-encode
component of the data share (up to a cross term suppressed by
$\sqrt{S_{\reenc}}$, bounded by the control's near-zero reading), so
$\|\Delta\bar r\|$ is read net of re-encode; the graph share and the
interaction pass through untouched, since the control has nothing to
cancel them with.

\paragraph{The drop bounds.} Both discarded terms obey exact bounds in
the retained shares (Cauchy--Schwarz on the expansion above),
\begin{equation}
|I_e| \;\le\; 2\sqrt{S_e\,\Phi_e}\,,
\qquad
\big|R'_e\big| \;\le\; 2\sqrt{\big(S_e + \Phi_e + I_e\big)\,\rho_e}
\;+\; \rho_e
\;+\; O\big(\kappa_\perp^2\kappa_\parallel,\; \kappa_\perp^4\big),
\label{eq:dropbounds}
\end{equation}
with $\rho_e = \|R_e^{\perp}\|^2 / (2\|\bar g_{\src}\|^2)$ the
remainder branch's own share. The first says the interaction can never
exceed the retained sum ($2\sqrt{S_e\Phi_e} \le S_e + \Phi_e$) and is
automatically negligible wherever one share dominates the other, so
assumption (ii) carries independent content only near share
crossover, which is precisely where Appendix~\ref{sec:headtohead} finds
it failing. The second says the first-order part of the remainder
enters at half power, one factor of $\sqrt{\rho_e}$ per retained
share, so assumption (i) reduces to the branch-level statement that
$R_e$ is small against $B_e$ and $C_e$, plus the small-$\kappa$
domain for the tail. The
control counterpart $R'_{\reenc}$ needs no bound of its own: the
control's degenerate expansion is itself the measured near-zero row of
Table~\ref{tab:floor}, which pins $R'_{\reenc}$ jointly with
$S_{\reenc}$.

\paragraph{What is not derived.} The loading
$\|\delta g^{\perp}\| = a_e\, \|\Delta\bar r(\sigma)\|$ (that the orthogonal
gradient mismatch is proportional to the residual mismatch with a
$\sigma$-independent route gain) is an empirical ingredient
(the data-branch factorization at the amplitude level), fit and tested
held-out in Appendix~\ref{sec:headtohead}. The geometry above fixes only how a
given $\delta g$ is read into gap units. What \emph{is} derivable from
first principles is only a one-sided envelope,
$\|\bar J^{\!\top}\Delta\bar r\| \le \sigma_{\max}(\bar J)\,
\|\Delta\bar r(\sigma)\|$, which would turn the data term into an
inequality in the observed mismatch norm given a $\sigma$-uniform bound
on the mean Jacobian's top
singular value. Promoting the envelope to a proportionality with a
$\sigma$-independent gain requires the mismatch \emph{direction}
$\widehat{\Delta\bar r}(\sigma)$ to hold a $\sigma$-stationary
alignment with $\bar J$'s singular directions (isotropy, for instance,
would give gain $\|\bar J\|_F/\sqrt{n}$) while $\bar J$ itself varies
with $\sigma$ through the network input; the image- and
route-specific mismatch directions of Appendix~\ref{app:instrument}
license no such axiom. Hence the loading stays an assumption with a
designated probe rather than a theorem.

\section{Instrument details and secondary reads}
\label{app:instrument}

\paragraph{Estimator.} Per image, arm, and $\sigma$-bin, adapter gradients
are accumulated over $D$ stratified draws and flattened; cosines are taken
between accumulated vectors. Per-bin cosines at $D{=}8$ are not comparable
\emph{across} bins (the floor drops where $\|g\|$ is small); the gap
subtraction against the same-bin floor is the valid read. Split-half
reliability over images of each bin-mean curve is mandatory
(observed $0.73$--$0.88$ on verdict runs); a per-image variant of the same
quantity has split-half reliability ${\approx}\,0$ and is not used.

\paragraph{Controls.} The re-encoding arm prices the VAE decode--encode
round trip that demotion necessarily pays; in pre-debiasing runs its gap
served as the validity band ($\pm 0.04$; observed $|\cdot| \le 0.054$
across all bins of the main runs, per-module-group $\le 0.045$ in the
split runs), a role superseded by the paired debiased read of
\S\ref{sec:instrument}. The redraw floor prices finite-draw noise.
Density-weighting the uniform bins by the trainer's $\sigma$-density
reproduces earlier $\sigma$-marginalized measurements from the same
instrument family (consistency check).

\paragraph{Debiasing, implementation.} Under
\texttt{--self\_floor}, every arm (re-encoding and each demote variant)
runs a second independent draw set from a disjoint seed stream, giving
per-arm self-cosines alongside the native floor; Eq.~\ref{eq:debias} is
then computed per image and bin, and the map read is the paired
excess $\gap_{e,i} = \dgap_{e,i} - \dgap_{\reenc,i}$ with non-finite
values (negative self-cosines under the square root at low $D$) and
$|\gap_{e,i}| > 1.5$ trimmed ($0$--$5$ of $40$ images per cell). Under
\texttt{--draw\_sweep}, draw prefixes are nested ($D = 64$ contains the
$D \le 32$ sets; prefix sums, no extra forwards) and the $c/D$ fit of
\S\ref{sec:instrument} is run on route means with a bootstrap CI
over images. Finite-draw cosines are kernel-path sensitive: twin runs
sharing a warm compiler kernel cache agree to $|\Delta\cos| \le 0.015$,
while a run compiled to a different kernel set lands up to
${\approx}\,0.3$ away at $D{=}2$, so gap/floor/self-floor pairings
are not compared across processes, a guarantee the single-run
instrument provides by construction; a \texttt{--deterministic} mode
(bit-exact across twin runs) covers the paired training A/Bs.

\paragraph{Finite-draw bias, measured.} Detail behind the magnitudes
quoted in \S\ref{sec:instrument}: the $c/D$ fit to the uncorrected
$1024{\to}896$ endpoint decay gives $c \approx 0.5$. The native
floor makes the point sharpest: at the verdict grid's $D{=}8$--$16$
the uncorrected endpoint self-cosine reads $0.79$--$0.85$, a naive
read pricing the native gradient's agreement with \emph{itself} at a
$0.15$--$0.2$ gap, before the nested sweep ($D = 4 \ldots 64$)
extrapolates it to the unit self-cosine reported in
\S\ref{sec:instrument}.

\paragraph{The U-shaped denominator (behind \S\ref{sec:ourscored}).}
The regimes of the total gradient norm $\|\bar g_{\src}(\sigma)\|$
($2.6 \to 0.4$ at $\sigma\!\approx\!0.3 \to 9.3$;
Fig.~\ref{fig:estimator}): large at high $\sigma$, where the model
pulls composition out of the prior; small in the mid-$\sigma$
refinement regime; inflated again at low $\sigma$ by irreducible
$\epsilon$. A monotone absolute mismatch over this U-shaped norm
yields the measured mid-$\sigma$-peaked curves: bin-mean gap
tracks $1/\|\bar g_{\src}\|$ (Spearman $+0.55$--$+0.90$). The
renormalization check reads the amplification
directly: multiplying each bin-mean gap by
$\|\bar g_{\src}(\sigma)\|$ flips the anomalous low-$\sigma$ dip into
the monotone-in-$\sigma$ maximum in every arm of both verdict runs,
which is the low-norm amplification behind the mid-$\sigma$ peak of the
measured curves. This shape read is post-hoc consistency analysis;
all safety criteria stay in cosine units, since direction is what a
normalized optimizer consumes.

\paragraph{Endpoint and x-zero modes.} The $\sigma{=}1$ endpoint bin feeds
input exactly $\epsilon$ (the input-mediated data term vanishes by
construction). The x-zero mode zeroes $x$ in input \emph{and} target on
every grid, keeping captions and exact demoted latent shapes; with the
$(1{-}\sigma)x$ term absent, low-$\sigma$ bins are off-manifold and the
$\sigma{=}1$ read is primary. In that regime the residual is
${\approx}\,{-}\hat x_{\mathrm{prior}}$, so ``graph-dominated'' includes
the model's grid-conditioned prior; the prior-distance probe
(\S\ref{sec:ourscored}), a forward-only comparison of the model's
caption-conditioned prior across grids, subsequently dissociated the
prior from the floor's route ordering: its route distances are flat
where the floors are strongly ordered, ruling out a
resolution-conditioned prior or a training-distribution discontinuity
as the carrier. In the target-strength ($\alpha$) sweep the
mid-sweep points $\alpha \in [0.5, 0.75]$ are unreadable by construction
and excluded: there the native residual $\alpha x - \hat x$ passes near
cancellation, the gradient norm dips ($60 \to 33$) and the redraw floor
falls to $0.87$, so the estimator reads through a small gradient norm:
the low-norm amplification above, surfacing along the $\alpha$ axis.
The well-conditioned anchors behind the $\alpha$-flatness claim of
\S\ref{sec:ourscored}: $768$ reads $+0.070 \pm 0.015$ at $\alpha{=}0$
vs $+0.049 \pm 0.010$ at $\alpha{=}1$ (control slope $+0.003$), and
$512$ reads $+0.269 \pm 0.041$ vs $+0.337 \pm 0.067$.

\paragraph{Parallel-landing decomposition of the target term (behind
\S\ref{sec:ourscored}).} The $\alpha$-flatness has an exact mechanism.
Because the objective is quadratic, $\bar g$ is affine in $\alpha$, so
the exact target-mediated gradient $t = \bar g(1) - \bar g(0)$ is
computable from the endpoint arms; measured, it is real and large
($\|t_{\src}\|/\|\bar g_{\src}\| \approx 2.2$, so $J^{\!\top}$ does not
annihilate target content), but demotion's change to it lands almost
entirely \emph{along} $\hat g_{\src}$ ($\kappa_\parallel =
-0.75/-1.18/-1.86$ against $\kappa_\perp = 0.09/0.14/0.20$, an
$8$--$9\times$ parallel dominance reproducible across draw sets). The
cosine gap is blind to a parallel rescaling (Eq.~\ref{eq:exact}) and
the orthogonal part enters only at second order
($\kappa_\perp^2/2 \approx 0.004$--$0.02$), so the $\alpha$-flat result
stands \emph{because} the target term lands parallel, not because it
vanishes. In this read the per-image orthogonal loading is
${\sim}10\times$ the aggregate's: image-specific directions cancel in
the mean, the same motif as the residual-direction read below. For
$896$ and $768$ the floors sit at or below the verdict grid's
resolution $\epsstar$, which the endpoint sweeps resolve only
through their order-of-magnitude larger draw budget. The statement
``the high-$\sigma$ plateau is the floor'' therefore sharpens to: the
floor is what the plateau converges to once the data term vanishes and
the estimator can resolve it.

\paragraph{Direction-resolved mismatch (behind
\S\ref{sec:ourscored}).} The route-uniformity of
$\|\Delta \bar r(\sigma)\|$ is a
property of the \emph{amplitude} only. Saving the per-image mismatch
vectors and comparing $\Delta \bar r$ directions with a split-half
attenuation correction, non-adjacent route pairs are near-orthogonal at
low $\sigma$ (corrected cosine $0.00$--$0.08$ at $\sigma \le 0.375$)
with only a weak shared component at high $\sigma$ ($+0.2$--$0.3$ at
$\sigma \ge 0.875$); the top mode of a stacked SVD carries
$0.33$--$0.36$ of the energy against the $0.25$ rank-4-uniform baseline
(a rank-one common mode would give ${\approx}\,1$); and cross-image
direction consistency is zero at every (route, $\sigma$) cell (max
$+0.019$ at split-half reliabilities $0.73$--$0.99$), so the mismatch
direction is fully image-specific, refuting a grid-conditional
composition prior (``small canvas $\Rightarrow$ portrait framing'') as
the carrier, at least under full captions. What \emph{is} shared is
scale and spectral drift: $\Delta \bar r$ energy migrates toward low
frequencies as $\sigma$ rises (low-third share $0.40 \to 0.67$ by
$\sigma = 0.875$), composition-scale but per-image. Eq.~\ref{eq:twoterm}
only ever consumes the norm, so the account is untouched; what sharpens
is the open question, now ``why is only the \emph{amplitude} universal
when the directions are image- and route-specific.'' A closed-form
candidate for cross-band coupling exists, namely the flow posterior
covariance, whose off-diagonal $D_z v^*$ entries are
$-\big((1{-}\sigma)/\sigma^3\big)\,\mathrm{Cov}(x_\omega, x_{\omega'}
\mid z)$ \citep{xing2026divunc} and are set to zero by the
diagonal-Gaussian premise; but its rank-one common-mode version is
excluded by the SVD read above.

\begin{figure}[t]
\centering
\includegraphics[width=0.55\linewidth]{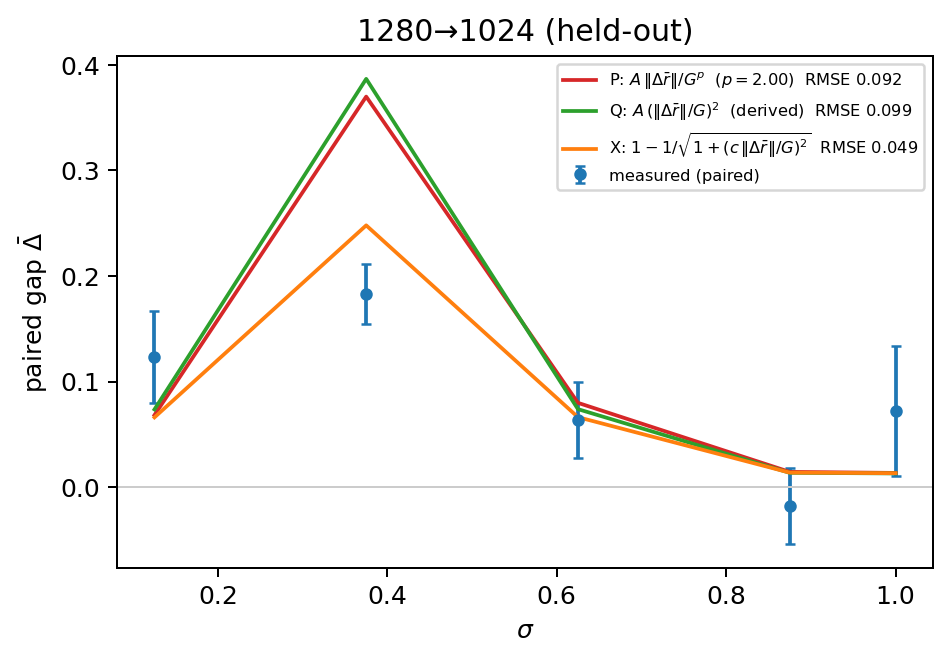}
\caption{The second held-out route, $1280{\to}1024$, under all three
functional forms for the data term of Appendix~\ref{sec:headtohead}
(fitted power, derived quadratic, exact angular link; legend
$G := \|\bar g_{\src}(\sigma)\|$), predicted from
governors fitted on other routes. This is the one route where the
forms visibly separate: the derived quadratic systematically
overshoots the mid-$\sigma$ peak; the exact angular link's saturation
removes the overshoot (RMSE $0.049$ vs ${\approx}\,0.095$), which is
why the exact link is the form displayed in Fig.~\ref{fig:accounts}.}
\label{fig:e51280}
\end{figure}

\paragraph{Held-out validation and refit, sources.} The fit consumes the
debiased paired curves of the $1024$-tier verdict run, the $1280$-tier
curves (pre-debiasing paired reads; see \S\ref{sec:limitations}),
the route-uniform mismatch norm $\|\Delta \bar r(\sigma)\|$ from the
residual probe, and
each run's own bin-mean gradient norm $\|\bar g_{\src}(\sigma)\|$ (gap
and norm always from the same process, per the kernel-path rule). The
shared-power fit scans $p \in [1, 2]$ jointly across fit routes with
per-route weighted least squares; the exact-link fit is a per-route grid
with profile-likelihood uncertainties on $a_e$. Oracle baselines are
quadratic-in-$\sigma$ fits on the held-out data itself (the noise
ceiling for a 3-dof curve); the spectral baseline on its committed
region is gap $= 0$ for $\sigma > \sigma_{\mathrm{eq}}$, and its
curve-level transport is the bridge of Fig.~\ref{fig:e83}.

\paragraph{The floor law's identifiability.} Two anchors and one
held-out check cannot identify a functional form, and the cosine gap is
the wrong scale to fit one in: it saturates (Eq.~\ref{eq:exact}).
Refitting the same anchors in the unsaturated perturbation-energy units
$\kappa_{\mathrm{eff}}^2 = (1-F)^{-2} - 1$ gives $\tau \approx 860$
tokens and predicts $F(2160) = +0.096$, equally consistent with the
measured $+0.092 \pm 0.012$; the same two anchors pushed through
$e^{-\sqrt{n}/\ell}$ or a power law $n^{-p}$ predict $+0.085$ and
$+0.075$. Every form falls within the union of the measured CI and the
prediction's own bootstrap spread, so the exponential is one member of
a family our operating points cannot distinguish, and the mechanical
reading of $\tau$ ($\approx 860$--$1040$ tokens across the two unit
conventions) as an effective-rank decay length of the early-block
gradient operator is a hypothesis, not a law. It is a discriminable
one: a spectral-tail account predicts a source-capacity dependence
$F \propto e^{-n/\tau} - e^{-n_0/\tau}$ where the
absolute-target-capacity governor predicts none (at our operating
points the secant correction sits within calibration slack, so the
existing anchors do not discriminate); the designated run is a
fixed-target, varied-source token ladder, not yet scheduled.

\paragraph{Bootstrap bands and the leave-$896$-out lane
(Fig.~\ref{fig:accounts}).} The bands are a full-pipeline image bootstrap
($B{=}1000$, fixed seed; $981$ draws kept): resample images with
replacement within each run ($N{=}40$ / $N{=}24$), re-bin the paired
curves, refit the exact-link $(a_e, \Phi_e)$ on every fit route,
re-derive the ratio governor and floor law, re-predict.
$\|\Delta \bar r(\sigma)\|$ and
the bin-mean $\|\bar g_{\src}(\sigma)\|$ are run-level instruments with
no per-image decomposition and are held fixed. The replica includes the
committed floor law's positive-floor filter ($\Phi_e > 0.005$): in
$41\%$ of draws the resampled $1120$ floor clears the filter and the law
re-anchors on the $(896, 1120)$ legs rather than $(512, 896)$. The
bands therefore contain this branch variability, which is the reason
\S\ref{sec:trainer} deploys the measured map rather than the predicted
one. The $896$ curve in Fig.~\ref{fig:accounts} is the post-hoc
leave-$896$-out lane under the exact link ($c$ from the ratio-twin
$1120$ alone $= 0.142$ vs $0.150$ fitted on $896$ itself; floor law
through $512{+}1120$, whose fragile $\Phi_{1120}$ leg enters the log
fit clamped positive ($36\%$ of bootstrap draws clamp), giving
$F(3012) = +0.012$ vs measured $+0.019$ $[+0.010, +0.030]$; curve RMSE
$0.073$).

\paragraph{Interventional $B/C$ ledger, implementation.} Routes
$1024{\to}\{896, 768, 512\}$, $\sigma \in [0.5, 1.0]$ in four bins plus
the $\sigma{=}1$ endpoint bin, $D{=}8$ draws per bin, $N{=}24$ images,
deterministic mode, two independent draw sets per arm. Each arm's
draw-summed adapter gradient is stored per bin; the interventional split
is $B = \bar g_{\mathrm{repromote}} - \bar g_{\src}$ (the data
intervention at fixed native graph: the demote--re-promote arm's input
passed through the downscale$\to$upscale$\to$encode pipeline but
evaluated on the native grid) and
$C = \bar g_{\mathrm{demote}} - \bar g_{\mathrm{repromote}}$ (the graph
intervention at fixed demoted data). The quadratic shares $S$, $F$, $I$
and $\rho = \cos(B^{\perp}, C^{\perp})$ use cross-draw-set inner
products to cancel shared draw noise; the debiased estimates are not
confined to $[-1, 1]$ (hence $\rho = -1.24$ at one low-amplitude
bin).
The exact counterfactual angles $h(B)$, $h(C)$, $h(B{+}C)$, with
$h(X) = 1 - \cos(\bar g_{\src}, \bar g_{\src} + X)$, are computed on
the raw arm means; since $\bar g_{\dem} = \bar g_{\src} + B + C$ by
construction, the realized demotion distance is exactly $h(B{+}C)$, so
each account's predicted-to-realized ratio is direct. The
no-interference scalar account $S{+}F$ overpredicts the realized
in-window gap $2.4$--$3.8\times$ at $768$ ($1.6$--$5.8\times$ at $896$,
$1.2$--$3.0\times$ at $512$); the fully additive counterfactual
$h(B) + h(C)$ overpredicts $3.5$--$8\times$, the realized $h(B{+}C)$
being only ${\sim}20$--$30\%$ of the additive sum in-window. Including
$I$ is what removes the overprediction; the vector account $B{+}C$ is
exact by construction. One caveat on the units of
Table~\ref{tab:e9ledger}: in-window
$|B^{\perp}|, |C^{\perp}| \approx 0.5$--$1.0\,\|\bar g_{\src}\|$,
outside the quadratic truncation's domain, so the truncated ledger
$S{+}F{+}I$ \emph{under}predicts realized magnitudes ${\sim}3$--$4\times$
(median ratio $0.24\times$). The ledger licenses sign, decomposition,
and window localization; gap magnitudes are quoted from $h(\cdot)$.
Finally, the re-encoding proxy check re-references the data leg against
the control arm, $B_{\reenc} = \bar g_{\mathrm{repromote}} -
\bar g_{\reenc}$: $|B^{\perp}_{\reenc}|$ agrees with $|B^{\perp}|$ to
${\sim}4\%$ at the signal-carrying low-$\sigma$ bins (worst $\pm 23\%$
only where $B$ is already small), so the shared downscale$\to$encode
pipeline cost is a minor share of the data intervention. This closes
the one open item on the floor ledger's first share, where the
re-encoding control (native decode$\to$re-encode) is a \emph{proxy}
for the pipeline cost demotion actually pays.

\begin{figure}[t]
\centering
\includegraphics[width=\linewidth]{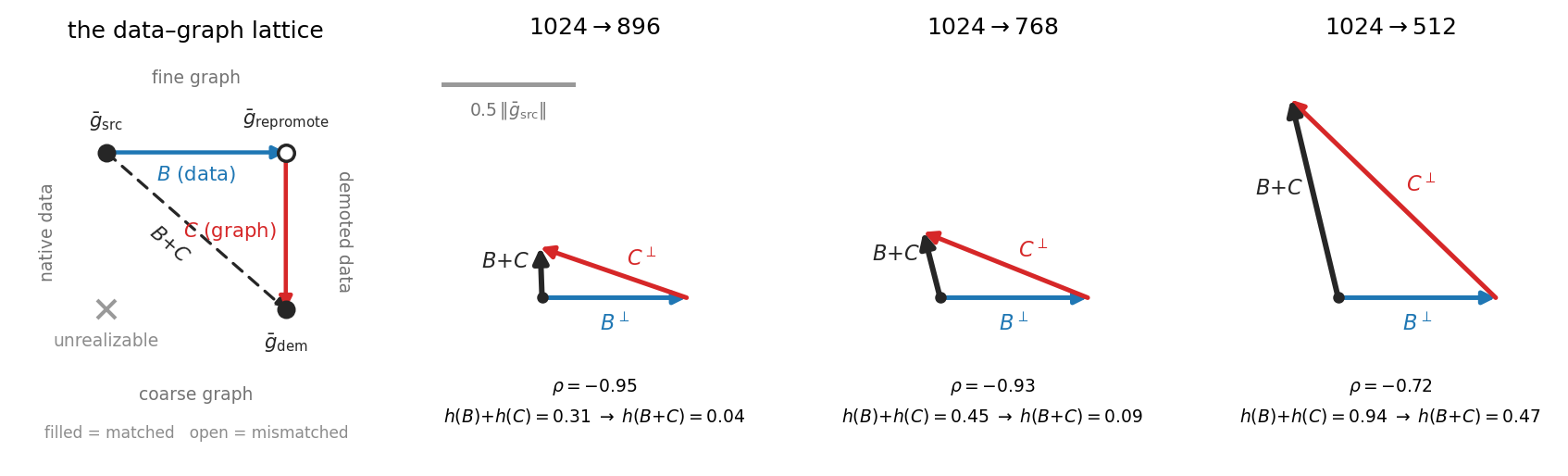}
\caption{\textbf{The ledger's geometry.} Left: the three arms as
corners of the data--graph lattice. $B$ and $C$ are the partial
differences along the lattice's only realizable path — the fourth
corner (native content on the coarse graph) does not exist — so the
split is a round trip off the matched configuration
($\bar g_{\src}$, $\bar g_{\dem}$) through the single mismatched
corner ($\bar g_{\mathrm{repromote}}$). Right: the measured legs
tip-to-tail at $\sigma = 0.563$ (Table~\ref{tab:e9ledger}), to a
common scale in units of $\|\bar g_{\src}\|$; the drawn angle is the
debiased $\rho$. Amplitude-matched legs leave only a small resultant
($896$, $768$); the $512$ graph leg is ${\sim}1.8\times$ the data
leg, and the unmatched share survives as the realized gap.}
\label{fig:ledgergeom}
\end{figure}

\paragraph{The ledger's geometry: why the legs anti-align.}
The near anti-parallelism of Table~\ref{tab:e9ledger} is structural
rather than incidental (Fig.~\ref{fig:ledgergeom}). The three arm means are values of one map on a
$2 \times 2$ data--graph lattice — $\bar g_{\src}$ at (native
content, fine graph), $\bar g_{\mathrm{repromote}}$ at (demoted
content, fine graph), $\bar g_{\dem}$ at (demoted content, coarse
graph) — all living in the shared adapter-parameter space, so the
differences are literal, and $B$ and $C$ are the partial differences
along the path (native, fine) $\to$ (demoted, fine) $\to$ (demoted,
coarse). That is the lattice's only realizable path: its fourth
corner, native content on the coarse graph, does not exist, since
placing native content on the coarse grid \emph{is} the demotion
pipeline — the coarse graph forces the demoted data. The endpoints
are the two content--graph \emph{matched} corners (native content on
its own grid; band-limited content on the grid whose Nyquist matches
it); the unavoidable waypoint is the single mismatched corner. The
split is therefore a round trip off the matched configuration and
back — $B$ creates the content--graph mismatch at fixed graph, $C$
removes it at fixed data — so, to leading order in the mismatch,
whatever share of each leg passes through it is the same vector with
opposite sign, which fixes the sign of $\rho$. Its depth is then
bookkeeping: projection off $\hat g_{\src}$ is linear, so
$|(B{+}C)^{\perp}|^2 = |B^{\perp}|^2 + |C^{\perp}|^2
+ 2\rho\,|B^{\perp}||C^{\perp}|$, and the ledger carries two
independent measurements — how far the waypoint sits from the corners
(the leg amplitudes) and how nearly the endpoints coincide (the
resultant) — with $\rho$ their law-of-cosines consequence; the
separately debiased sign column is a consistency check on the raw
$h(\cdot)$ reads, not a third fact. Two corollaries follow, and the
data confirms both. Route verdicts reduce to amplitude matching, the
realized $h(B{+}C)$ collapsing where
$|B^{\perp}| \approx |C^{\perp}|$ (the window-center localization of
Appendix~\ref{sec:headtohead}); and what no matching can cancel is
the graph-intrinsic share of $C$ with no data-side counterpart — the
resultant collects exactly the residue the floor ledger below
decomposes into $\mathrm{RoPE}_e + \mathrm{Resid}_e$, and at the
$\sigma = 1$ endpoint, where the input is $\epsilon$ on every arm and
the data leg has nothing left to act on, that uncancelled residue is
the whole of the floor.

\paragraph{Floor decomposition, detail (behind
\S\ref{sec:ourscored}).} Splitting the probe gradients per module type
($15$ groups, including separate rows of the fused QKV projections) and
per block ($28$) localizes the floor in \emph{depth}, not module type:
early blocks (0--9, peaking at 3--8) carry ${\sim}3\times$ the
late-block gap, uniformly across every module type within a block (at
$512/\sigma{=}1$ every type sits in $0.22$--$0.31$;
Table~\ref{tab:depth}). The mechanism picture: the first ${\sim}10$
blocks build grid-calibrated token statistics, the divergence
propagates into every parameter type's gradient in those blocks roughly
equally, and deeper blocks inherit a washed-out version. Query and key
projections show \emph{zero} excess over value projections, which we
initially read as refuting a rotary mechanism. That was an instructive
error, since landing-side uniformity cannot localize \emph{origin}: a
perturbation originating in the position embedding propagates through
the block and lands on all module types. Hence the origin-side
positional-interpolation intervention (Table~\ref{tab:pi}; the paired
$\Delta$s are same-grid contrasts at matched draws, in which the
finite-draw bias cancels to first order). The $896$ route's small floor
sits below that probe's paired resolution and remains undecomposed. A
$\sigma$-resolved follow-up forecloses the route this suggests
(training $1024{\to}768$ through interpolated coordinates): with image
content in the input the stretched forward is off-manifold, and the
interpolated arm is \emph{worse} than the plain demoted arm through
$\sigma \in [0.56, 0.81]$ (paired $-0.05$ to $-0.11$), winning only at
$\sigma \ge 0.94$; the $768$ route's data term is in any case fatal on
its own ($+0.09$--$0.22$ over control across the window). On the
non-positional residue, the attention-over-$N$ piece has inference-side
prior art: attention-entropy corrections derive length-dependent
temperatures to preserve the $\log N$ entropy term
\citep{li2025infoscale}, and TIDE applies the same dilution correction
jointly to resolution and diffusion timestep \citep{liu2026tide},
the closest prior work to a $\sigma$-gated resolution correction,
though on the inference side; our gate is training-side and set by
measured gradient safety, not by an entropy invariance.

\paragraph{The floor as a ledger.} Combining the designated probes of
\S\ref{sec:ourscored}, the floor decomposes additively, each share
measured by its own intervention:
\begin{equation*}
\Phi_e \;=\;
\underbrace{\reenc}_{{\approx}\,0 \text{ by control}}
\;+\; \underbrace{\text{target content}}_{\text{lands }\parallel\;
\hat g_{\src}\text{: angular share}\,{\approx}\,0}
\;+\; \underbrace{\mathrm{RoPE}_e}_{\text{erased by PI at }\sigma{=}1}
\;+\; \underbrace{\mathrm{Resid}_e}_{\text{remainder}}
\end{equation*}
with totals $0.02$--$0.04$ ($896$; below the PI probe's paired
resolution, undecomposed), ${\approx}\,0.07$ ($768$; $\mathrm{RoPE}$
the large majority, $\mathrm{Resid} \approx 0$), and ${\approx}\,0.30$
($512$; $\mathrm{RoPE} \approx 0.10$, $\mathrm{Resid} \approx 0.20$).

\begin{table}[t]
\caption{The interventional $B/C$ ledger behind
Appendix~\ref{sec:headtohead}: per (route, $\sigma$-bin) orthogonal
amplitudes (units of $\|\bar g_{\src}\|$), interaction geometry, the
cross-set-debiased quadratic shares, and the exact counterfactual
angles. In-window $S{+}F{+}I$ is sign- and localization-accurate but
underpredicts $h(B{+}C)$ (small-perturbation truncation out of
domain); magnitude claims in the text use $h(\cdot)$.}
\label{tab:e9ledger}
\begin{center}
\small
\begin{tabular}{llrrrrrrrrr}
\toprule
route & $\sigma$ & $|B^{\perp}|$ & $|C^{\perp}|$ & $\rho$ & $S$ & $F$ & $I$ & $h(B)$ & $h(C)$ & $h(B{+}C)$ \\
\midrule
$896$ & $0.563$ & $0.55$ & $0.59$ & $-0.95$ & $0.103$ & $0.115$ & $-0.206$ & $0.107$ & $0.206$ & $0.037$ \\
      & $0.688$ & $0.42$ & $0.39$ & $-0.91$ & $0.036$ & $0.059$ & $-0.085$ & $0.083$ & $0.077$ & $0.062$ \\
      & $0.813$ & $0.20$ & $0.23$ & $-0.94$ & $0.016$ & $0.019$ & $-0.033$ & $0.022$ & $0.027$ & $0.010$ \\
      & $0.938$ & $0.10$ & $0.10$ & $-1.24$ & $0.002$ & $0.003$ & $-0.006$ & $0.007$ & $0.006$ & $0.003$ \\
      & $1.000$ & $0.05$ & $0.06$ & $-0.71$ & $0.001$ & $0.001$ & $-0.001$ & $0.002$ & $0.003$ & $0.002$ \\
\midrule
$768$ & $0.563$ & $0.57$ & $0.68$ & $-0.93$ & $0.139$ & $0.200$ & $-0.309$ & $0.107$ & $0.346$ & $0.088$ \\
      & $0.688$ & $0.56$ & $0.57$ & $-0.93$ & $0.100$ & $0.136$ & $-0.218$ & $0.106$ & $0.242$ & $0.098$ \\
      & $0.813$ & $0.35$ & $0.42$ & $-0.93$ & $0.049$ & $0.067$ & $-0.105$ & $0.059$ & $0.110$ & $0.035$ \\
      & $0.938$ & $0.14$ & $0.17$ & $-1.02$ & $0.005$ & $0.009$ & $-0.014$ & $0.010$ & $0.033$ & $0.008$ \\
      & $1.000$ & $0.07$ & $0.09$ & $-0.62$ & $0.001$ & $0.004$ & $-0.003$ & $0.003$ & $0.014$ & $0.013$ \\
\midrule
$512$ & $0.563$ & $0.60$ & $1.09$ & $-0.72$ & $0.166$ & $0.403$ & $-0.372$ & $0.085$ & $0.854$ & $0.466$ \\
      & $0.688$ & $0.63$ & $0.88$ & $-0.86$ & $0.153$ & $0.332$ & $-0.388$ & $0.107$ & $0.623$ & $0.305$ \\
      & $0.813$ & $0.57$ & $0.72$ & $-0.87$ & $0.152$ & $0.226$ & $-0.323$ & $0.131$ & $0.285$ & $0.126$ \\
      & $0.938$ & $0.25$ & $0.29$ & $-0.96$ & $0.026$ & $0.036$ & $-0.058$ & $0.028$ & $0.201$ & $0.037$ \\
      & $1.000$ & $0.13$ & $0.15$ & $-0.88$ & $0.007$ & $0.010$ & $-0.015$ & $0.007$ & $0.379$ & $0.098$ \\
\bottomrule
\end{tabular}
\end{center}
\end{table}

\section{Head to head: predicting held-out routes}
\label{sec:headtohead}

\begin{table}[t]
\caption{\textbf{Head to head} on the same measured curves: RMSE in
cosine units over all $\sigma$-bins. The spectral column is the family
transported through our bridge (Fig.~\ref{fig:e83}) and is
$\delta$-inert. Our two columns are the derived small-perturbation form
(Eq.~\ref{eq:twoterm}) and the exact angular link (Eq.~\ref{eq:exact}),
both with \emph{zero per-route freedom} on held-out routes: amplitudes
and floors come from the two governors of \S\ref{sec:ourscored} fitted
on $\{1024{\to}896,\ 1024{\to}512,\ 1280{\to}1120\}$. The oracle is a
quadratic-in-$\sigma$ fit on the held-out data itself, the noise
ceiling for a 3-dof curve, so beating it on $1024{\to}768$ means the
prediction is at the data's own resolution.}
\label{tab:headtohead}
\centering
\small
\begin{tabular}{llcccc}
\toprule
route & status & spectral & two-term & exact link & oracle \\
\midrule
$1024{\to}768$ & held out & $0.147$ & $0.093$ & $0.093$ & $0.105$ \\
$1280{\to}1024$ & held out & --- & $0.092$ & $\mathbf{0.049}$ & $0.056$ \\
$1024{\to}512$ & fit route & $0.355$ & $0.093$ & --- & --- \\
\bottomrule
\end{tabular}
\end{table}

The probes of \S\ref{sec:scored} test each account's \emph{structure}. The remaining
test is predictive: fit the two-term reduction on three routes, then
predict two held-out routes from the governors alone. Fit routes:
$\{1024{\to}896,\ 1024{\to}512,\ 1280{\to}1120\}$; held out:
$\{1024{\to}768,\ 1280{\to}1024\}$. Per fit route the excess curve
$\bgap_e(\sigma)$ is fit with a route amplitude and floor on the
measured route-uniform $\|\Delta \bar r(\sigma)\|$ and each run's own
$\|\bar g_{\src}(\sigma)\|$; two governor models ($A(\mathrm{ratio})$
interpolating the fitted amplitudes, and an exponential floor law
$F(n) = F_0 e^{-n/\tau}$ in target tokens $n$) then generate the
held-out predictions with zero per-route freedom. Pass criteria were
pre-registered in the analysis script before the numbers.

\paragraph{Held-out results, and the comparison.}
All four pre-registered gates pass (Table~\ref{tab:headtohead},
Fig.~\ref{fig:accounts}). (i)~Held-out $1024{\to}768$: RMSE $0.093$,
\emph{better than an oracle quadratic fit on the held-out data itself}
($0.105$), and better than the spectral account both across all bins
($0.147$) and on the spectral account's own committed region, where it
predicts gap $0$ ($0.164$ against $0.096$ for ours).
(ii)~Held-out $1280{\to}1024$: RMSE $0.092$ under the derived quadratic
and $0.049$ under the exact link, against oracle $0.056$; inside the
pre-registered $2\times$ gate either way. (iii)~The ratio governor
holds at the amplitude level: the two ratio-$0.875$ routes agree at
$z = 0.14$ despite a $1.6\times$ difference in target capacity.
(iv)~The floor law $F(n) = 0.70\,e^{-n/1041\,\mathrm{tok}}$, fit on the
$512$ and $896$ floors, predicts $F(4825) = +0.007$ for the $1120$ grid
(fitted $+0.002$) and $F(2160) = +0.088$ for the held-out $768$ grid,
against a measured endpoint floor of $+0.092 \pm 0.012$; a post-hoc
leave-$896$-out re-derivation from $512{+}1120$ alone predicts the
$896$ floor at $+0.019$ under the fitted-power form ($+0.012$ under
the exact link, Appendix~\ref{app:instrument}) against the measured
$+0.019$ $[+0.010, +0.030]$. The floor law is thus correct on both floors
outside its fit set, though its point estimates vary with the draw: a
full-pipeline bootstrap puts $F(2160)$ at $68\%$ $[+0.042, +0.091]$,
so the licensed read is agreement, not exactness.

The margin in Table~\ref{tab:headtohead} has a structural reading, not
merely a numerical one. The spectral family's error is not a bad
tolerance, since $\delta$ is inert, but a missing term and a missing
governor: it has nothing to put in the high-$\sigma$ region where the
floor lives, and nothing to distinguish $1280{\to}1024$ from a
same-ratio route at a quarter of the token count. Our account's
held-out accuracy comes from precisely the two ingredients it adds.

\paragraph{Which functional form, and what the fit does not identify.}
Refitting the same pipeline under three forms for the data term (a
fitted power $A\,m/G^p$ with $p$ shared and free, the derived
quadratic, and the exact link) resolves the form question in the
geometry's favor twice over. The free exponent's own optimum lands on
$p = 2.00$ exactly, recovering the derived quadratic on its own; and
the exact link is the best held-out predictor (mean RMSE $0.071$ vs
$0.093$--$0.094$), with its entire margin coming from $1280{\to}1024$,
where the quadratic systematically \emph{overshoots} the mid-$\sigma$
peak and the exact link's saturation removes the overshoot (Appendix
Fig.~\ref{fig:e51280}). The governors are form-invariant. Two things
the fit does \emph{not} establish: the prediction is not within
the verdict resolution anywhere ($\chi^2$/bin $7.6$--$9.2$ held-out
under the quadratic), so the licensed claim is shape, magnitude class,
and governors at ${\sim}0.07$--$0.09$ RMSE; and the floor law's
functional form is unidentified at our operating points
(the identifiability analysis is in Appendix~\ref{app:instrument}).

\paragraph{The reduction's domain: one signed failure, and its
mechanism.}
The $768$ route's mid-$\sigma$ window is the reduction's one structural
failure, and its shape is diagnostic. Measured excess sits at zero
across $\sigma \in [0.56, 0.94]$, \emph{below} the predicted
$\Phi_{768} \approx 0.09$ and outside the bootstrap $95\%$ band
(Fig.~\ref{fig:accounts}): a positive two-term reduction cannot produce a gap
under its own floor, under any coefficients. In the four-term expansion
(Eq.~\ref{eq:fourterm}) this admits exactly two mechanisms, both
pre-registered ahead of the probe: either the projected interaction
is negative in the window ($I_{768} < 0$, in which case amplitude
matching predicts the window center sits where the two perturbation
legs have equal orthogonal magnitude, a testable localization), or the
graph share is itself $\sigma$-dependent, rewriting assumption (iii)
rather than (ii). The designated probe is vector-resolved: a
demote--re-promote arm reads the shares per (route, bin) from the
interventional split $B = \bar g_{\mathrm{repromote}} - \bar g_{\src}$,
$C = \bar g_{\mathrm{demote}} - \bar g_{\mathrm{repromote}}$
(Appendix~\ref{app:instrument}, Table~\ref{tab:e9ledger}).

When the probe is run, the data select the first branch, with its
localization confirmed:
$I_{768}(\sigma) < 0$ at every bin ($-0.31$ at $\sigma = 0.56$,
shrinking to $-0.014$ by $0.94$), with the two legs near anti-parallel
in-window ($\rho = \cos(B^{\perp}, C^{\perp}) \approx -0.93$), and the
predicted window center lands at $\sigma \approx 0.69$
($|B^{\perp}|/|C^{\perp}| = 0.98$ there), inside the measured
window. Branch (ii) is ruled out: $\Phi_{768}(\sigma)$ decreases
monotonically to its endpoint value and never drops below it, so the
endpoint reads of \S\ref{sec:ourscored} are untouched. The
anti-parallel geometry turns out to be universal
($\rho \in [-1.24, -0.62]$ across every route and bin), as the
round-trip reading of the split says it must be
(Appendix~\ref{app:instrument}, \emph{the ledger's geometry}), so what
distinguishes routes is amplitude matching, not sign; measured against
the exact counterfactual angles, roughly three quarters of the
additively-predicted gap is erased by the interference in-window
(Appendix~\ref{app:instrument}). The window therefore \emph{delimits
the reduction's domain}, and the derived four-term form represents the
failure rather than excusing it.

\section{The spectral account: spectrum and instantiation}
\label{app:spectral}

\begin{table}[t]
\caption{The spectral account (Eq.~\ref{eq:spd}) instantiated on the
measured latent spectrum ($P$ at the $896/768/512$ cuts
$= 0.025/0.029/0.065$; RAPSD in Fig.~\ref{fig:spectrum}):
predicted safe boundaries $t^{*}$ per route as the tolerance $\delta$
sweeps its range. $\delta_{\reenc}$ is the measured re-encode noise
floor, a zero-free-parameter anchor fixed by our own pipeline. No row
reproduces the measured pattern (last line): the $\delta$ that matches
the safe route's boundary predicts both failing routes safe by
$\sigma = 0.63$, and no $\delta$ can produce a floor.}
\label{tab:null}
\centering
\small
\setlength{\tabcolsep}{4pt}
\begin{tabular}{lcccl}
\toprule
tolerance $\delta$ & $1024{\to}896$ & $1024{\to}768$ & $1024{\to}512$ & \\
\midrule
$(1+P_\omega)/2$ (equal power) & $0.14$ & $0.15$ & $0.20$ & classical crossover \\
$0.025$ & $0.50$ & $0.52$ & $0.62$ & tuned to the measured $896$ gate \\
$0.01$ & $0.61$ & $0.63$ & $0.72$ & default of \citet{xiao2026spd} \\
$\delta_{\reenc} \approx 10^{-5}$ & $0.98$ & $0.98$ & $0.99$ & the parameter-free anchor \\
\midrule
\textbf{measured, debiased} & ${\approx}\,0.5$ &
none safe & unsafe ($8/9$ bins) & floors $.02$--$.04$ / $.06$--$.09$ / ${\approx}\,.3$ \\
\bottomrule
\end{tabular}
\end{table}

\begin{figure}[t]
\centering
\begin{subfigure}[t]{0.485\textwidth}
\includegraphics[width=\linewidth]{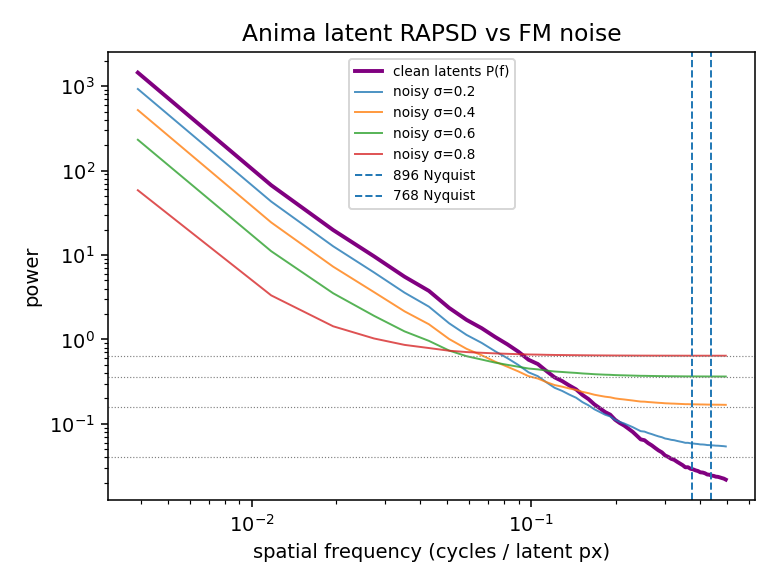}
\caption{Latent RAPSD against the flow-matching noise floor at four
noise levels; dashed lines mark the demoted grids' Nyquist cuts.}
\label{fig:rapsd}
\end{subfigure}\hfill
\begin{subfigure}[t]{0.49\textwidth}
\includegraphics[width=\linewidth]{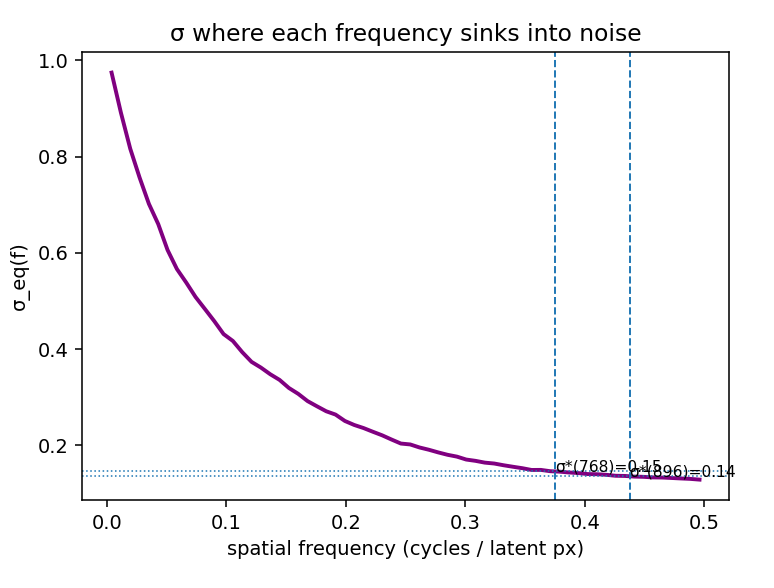}
\caption{The equal-power slice $\sigma_{\mathrm{eq}}(f)$ of the spectral
family. Predicted $\sigmastar$: $0.136$ / $0.146$ / ${\sim}0.20$ for
demotion to 896 / 768 / 512.}
\label{fig:sigmaeq}
\end{subfigure}
\caption{The spectral account, computed from the measured RAPSD and
transported to the training question (\S\ref{sec:spectral}): safety
would be predicted from $\sigma \approx 0.14$. Against the measurement
(Fig.~\ref{fig:verdict}) no tolerance reproduces the pattern
(Table~\ref{tab:null}): the equal-power crossover is off by
${\sim}3.5\times$, and the $2\times$ downscale, predicted safe by every
tolerance, is safe at no noise level.}
\label{fig:spectrum}
\end{figure}

\begin{figure}[t]
\centering
\includegraphics[width=\linewidth]{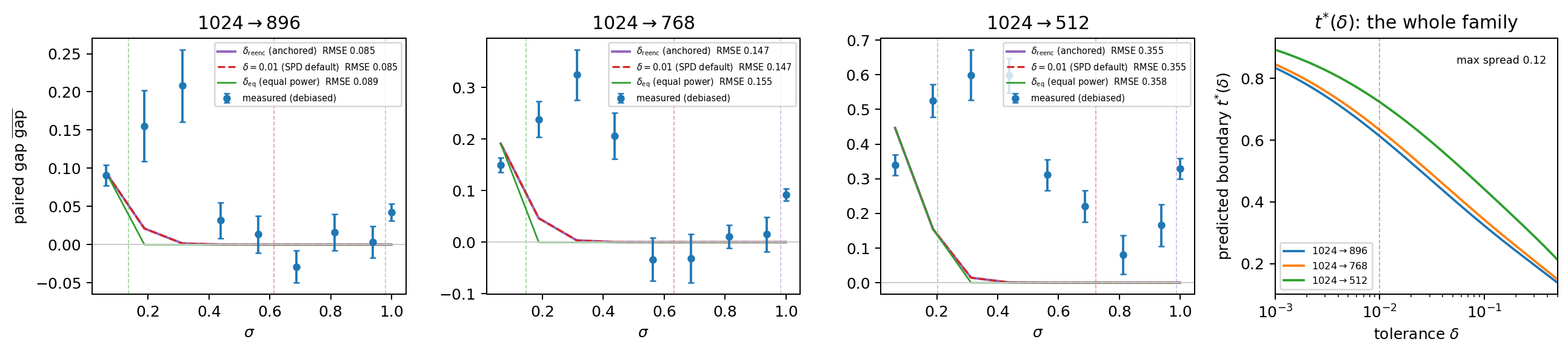}
\caption{The \emph{whole} tolerance family against the measurement:
the detail behind the single red curve of Fig.~\ref{fig:accounts}. Left
three panels: the measured debiased gap curves against the spectral
account transported through our bridge (the diagonal model's
destroyed-band mean-residual mismatch on the measured spectrum, through
the measured $\|\bar g_{\src}(\sigma)\|$ with a single gain calibrated
on the one safe route and no floor, which the family cannot express),
gated at $t^{*}(\delta)$ (dashed verticals) for the equal-power slice,
SPD's default $\delta = 0.01$, and the measured anchor
$\delta_{\reenc}$. The transported curve decays to zero by
$\sigma \approx 0.35$
on every route, so the two larger gates ($0.61$--$0.72$ and
$0.98$--$0.99$) act on an already-zero curve and coincide exactly;
the equal-power gate ($0.14$--$0.20$) is the only one that truncates
a nonzero curve, and it moves RMSE by at most $0.008$. All three therefore
land within $0.01$ of each other: $\delta$ is inert at the curve
level. Right: the boundary family $t^{*}(\delta)$; maximum route spread
$0.125$ in $\sigma$ over $\delta \in [10^{-3}, 0.5]$.}
\label{fig:e83}
\end{figure}

The tolerance $\delta$ of Eq.~\ref{eq:spd} is the account's one free
parameter, and this appendix instantiates the whole family. The choice
$\delta = (1+P_\omega)/2$ recovers the often-quoted equal-power
(input-SNR) crossover
$\sigma_{\mathrm{eq}} = \sqrt{P_\omega}/(1+\sqrt{P_\omega})$ as one
slice; \citet{xiao2026spd} provide an inference-side default
$\delta = 0.01$; and our own pipeline fixes a zero-free-parameter
anchor $\delta_{\reenc}$ by construction (below).

Table~\ref{tab:null} evaluates Eq.~\ref{eq:spd} at the demoted grids'
Nyquist frequencies $f_{\mathrm{cut}} = \tfrac{1}{2}\,e/e_0$ on the
$N{=}40$ mean radially-averaged latent power spectrum ($64$ radial
bins; $P$ at the $896/768/512$ cuts $= 0.025/0.029/0.065$;
Fig.~\ref{fig:spectrum}). Across
$\delta \in [0.001, 0.5]$ the mildest--harshest spread remains below
$0.13$, and the $1024{\to}896$ versus $896{\to}768$ boundaries (cut
frequencies $0.4375$ vs $0.4286$) separate by no more than $0.01$.
The parameter-free anchor $\delta_{\reenc}$ of \S\ref{sec:specscored} is
measured by the same estimator: RAPSD of the re-encode arm's latent error
(fresh resize--encode minus cached latent, the gradient probe's own
control chain) on the same probe set, interpolated at each route's cut
frequency. Measured ($N{=}40$, $64$ radial bins): error variance
$2.7\times10^{-5}$ against latent variance $0.419$;
$D(f_{\mathrm{cut}}) \approx 1.0\times10^{-5}$ at all three cuts
(p10--p90 over images within $[1, 2]\times10^{-5}$; per-band SNR
$P(f_{\mathrm{cut}})/D(f_{\mathrm{cut}}) \approx 2.5$--$4.4\times10^{3}$),
giving $t^{*}_{\reenc} = 0.98 / 0.98 / 0.99$ for $896/768/512$.

\paragraph{The ``wrong $\delta$'' objection, closed by construction.}
One might object that the right $\delta$ has simply not been chosen:
SPD's is an inference-side value selected on a speed--quality Pareto
ablation over generated images \citep{xiao2026spd}, and nothing ties it
to gradients. Our setting fixes one \emph{by construction}. The
demotion recipe pays a resize--VAE-encode round trip whose latent error
has measurable per-band power $D(f)$, expressed in
Eq.~\ref{eq:spd}'s own units, so $\delta_{\reenc}(e) :=
D(f_{\mathrm{cut}}(e))$ anchors the family with zero free parameters.
Measured (above), that anchor is tiny and image-generic, and the
boundary moves it forces are measured in Table~\ref{tab:null} and
Fig.~\ref{fig:e83}: every boundary is pushed to $t^{*} \approx 0.98$,
maximally conservative, and still structurally wrong in both
directions at once: it moves all three boundaries together (spread
$< 0.01$), predicts the measured-safe route safe only above
$\sigma \approx 0.98$ where the measurement shows it safe from $0.5$,
and, like every member of the family, predicts the $2\times$ downscale
eventually safe when no noise level is. The confrontation of
\S\ref{sec:specscored} therefore does not hinge on the choice of
tolerance: the structural failures of \S\ref{sec:spectral}(a--c) are
$\delta$-independent at the boundary level, and $\delta$ is inert
outright at the curve level (Fig.~\ref{fig:e83}).

\section{Controlled-adapter replication of the verdict map}
\label{app:e7}

\begin{figure}[t]
\centering
\begin{subfigure}[t]{0.49\textwidth}
\includegraphics[width=\linewidth]{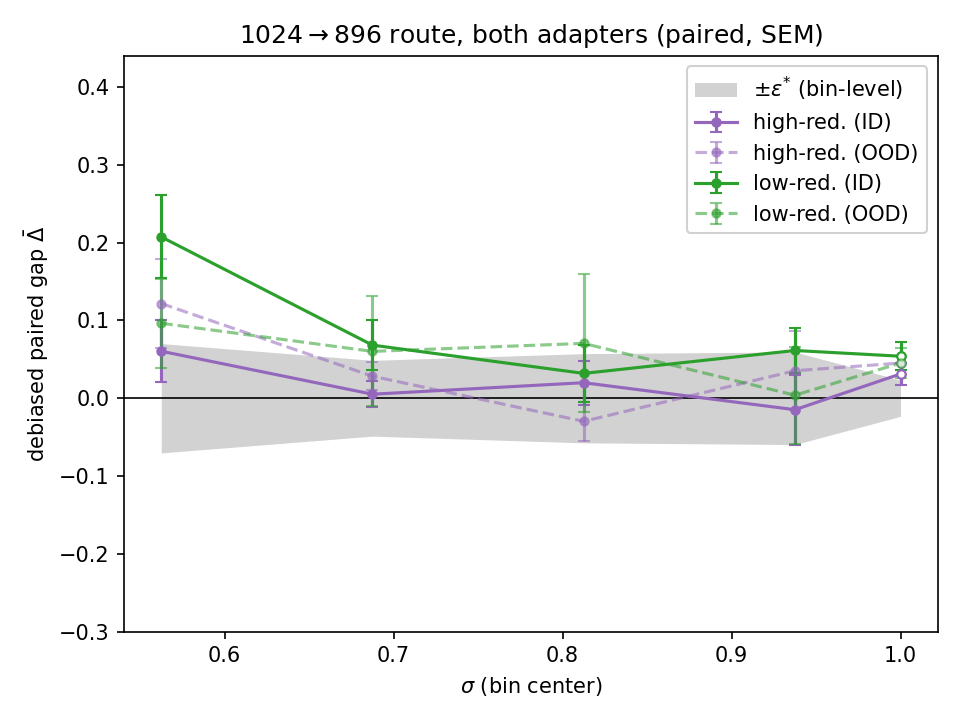}
\caption{$1024{\to}896$ (the safe route).}
\label{fig:e7route896}
\end{subfigure}\hfill
\begin{subfigure}[t]{0.49\textwidth}
\includegraphics[width=\linewidth]{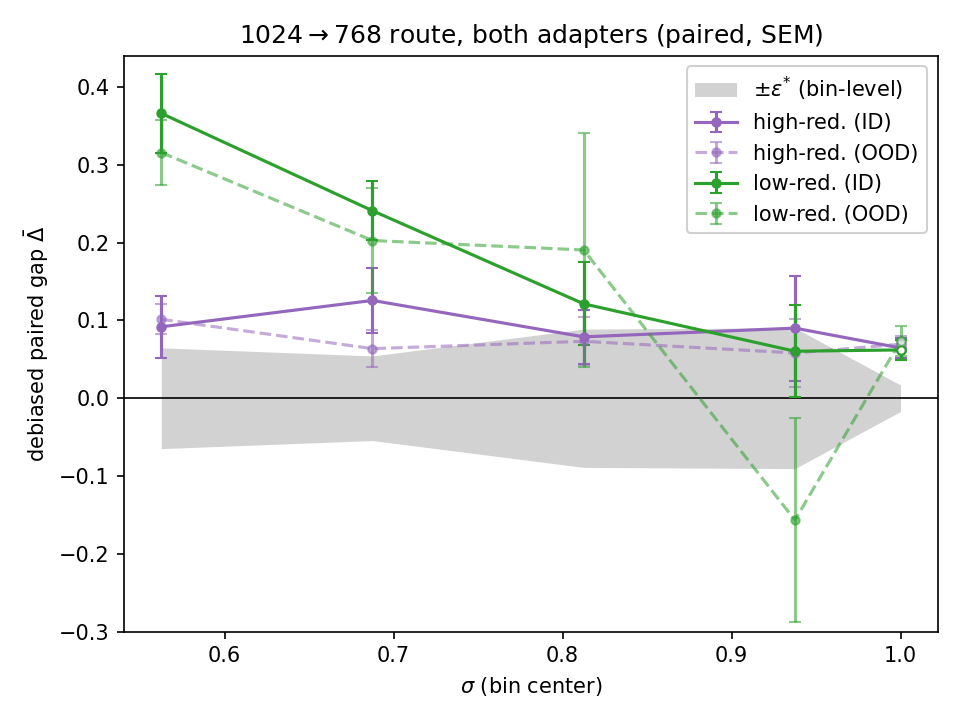}
\caption{$1024{\to}768$.}
\label{fig:e7route768}
\end{subfigure}
\caption{The Fig.~\ref{fig:gapcurves} recipe repeated on two controlled
adapters trained on opposite style clusters, one panel per route
($N{=}48$ stems per adapter, $D{=}8$/bin, high-$\sigma$ window; same
paired debiased estimator and trim; open markers the $\sigma{=}1$
endpoint bin; axes shared across panels). Unlike the main-text figures,
color encodes the \emph{adapter} (high- vs low-redundancy cluster);
solid curves are in-distribution stems, the adapter's own style
cluster ($n{=}36$: trained-on, held-out, and unseen-artist cells),
and dashed curves are out-of-distribution stems from the opposite
cluster ($n{=}12$). Gray band: bin-level $\pm\epsstar$ per
Eq.~\ref{eq:safe}, median of the two runs' full-$N$ SEMs for the
panel's route. The map replicates: on the safe route all four
adapter$\times$distribution curves reach the band inside the window,
on $768$ all four sit above it until the top, and within each panel the
dashed OOD curves track their solid ID counterparts.}
\label{fig:e7maps}
\end{figure}

The one-adapter limitation of \S\ref{sec:limitations} pre-registered a
controlled $2{\times}2$ factorial: two plain-LoRA checkpoints trained
under the original probe adapter's frozen recipe on opposite style
clusters (the top and bottom $12$ artists by median latent
redundancy, hereafter the high-redundancy, visually flat, and
low-redundancy, heavily textured, clusters), with stem-level
membership manifests frozen
before training ($124$ training images each, identical seeds). Each
adapter was probed on $48$ stems spanning four membership cells
($N{=}12$ each): trained-on, held-out images of trained artists,
unseen artists of the same cluster, and unseen artists of the
opposite cluster, with the cross-cluster stems shared between the two
runs so cross-adapter contrasts read paired per stem.
Figure~\ref{fig:e7maps} repeats the verdict-map recipe per route,
overlaying both adapters.

Three reads, stated at the bin-mean level (per-cell tables reproduce
from the archived per-image rows). First, the $(\text{route}, \sigma)$
shape replicates on both checkpoints (Fig.~\ref{fig:e7maps}). Second,
the factorial contrasts are null at instrument resolution: the
probe-style main effect and the adapter$\times$probe-style interaction
(the pre-registered verdict quantity) are both bounded below the
resolvable bin-mean effect (raw-gap paired interaction
$-0.022 \pm 0.027$ at $896$, $-0.015 \pm 0.059$ at $768$), and the
membership contrast (trained-on versus held-out) does not replicate in
sign across the two adapters. The map is therefore adapter- and
content-agnostic on this model at our resolution, supporting a
calibrate-once-per-model reading. Third, the one quantity that is
\emph{not} checkpoint-invariant is the absolute redraw-floor level:
in-window floor cosines sit at ${\approx}0.73$ for the
high-redundancy-cluster adapter versus ${\approx}0.50$ for the
low-redundancy-cluster adapter, uniformly across all four membership
cells. The floor's level is a property of the checkpoint (its gradient
stochasticity), while the safety map built on top of it is not.

\section{Render prompt for the arms figure}
\label{app:armsprompt}

The renders of Fig.~\ref{fig:kaaiyuki} share one positive prompt, a
held-out corpus-B caption, rendered at $1024^2$ with matched noise
seed across arms:
\begin{quote}
\scriptsize\texttt{safe, 2girls, kaai yuki, otomachi una, vocaloid,
zako (vocaloid), @channel (caststation), bag, black hair, black
skirt, blush stickers, bow, bowtie, collared shirt, diagonal-striped
bow, diagonal-striped bowtie, diagonal-striped clothes,
diagonal-striped neckerchief, emphasis lines, hair bobbles, hair
ornament, hands on own cheeks, hands on own face, holding, holding
phone, jitome, long sleeves, multiple girls, multiple views,
neckerchief, notice lines, one eye closed, outstretched arm, phone,
pink background, pink jacket, plaid clothes, plaid skirt, school bag,
small sweatdrop, striped bow, striped bowtie, striped clothes,
striped neckerchief, tongue, tongue out, v, v over eye, v over mouth,
white shirt, wink star}
\end{quote}

\section{Per-bin verdict numbers, and the historical record}
\label{app:tables}

\begin{figure}[t]
\centering
\begin{subfigure}[t]{0.46\textwidth}
\includegraphics[width=\linewidth]{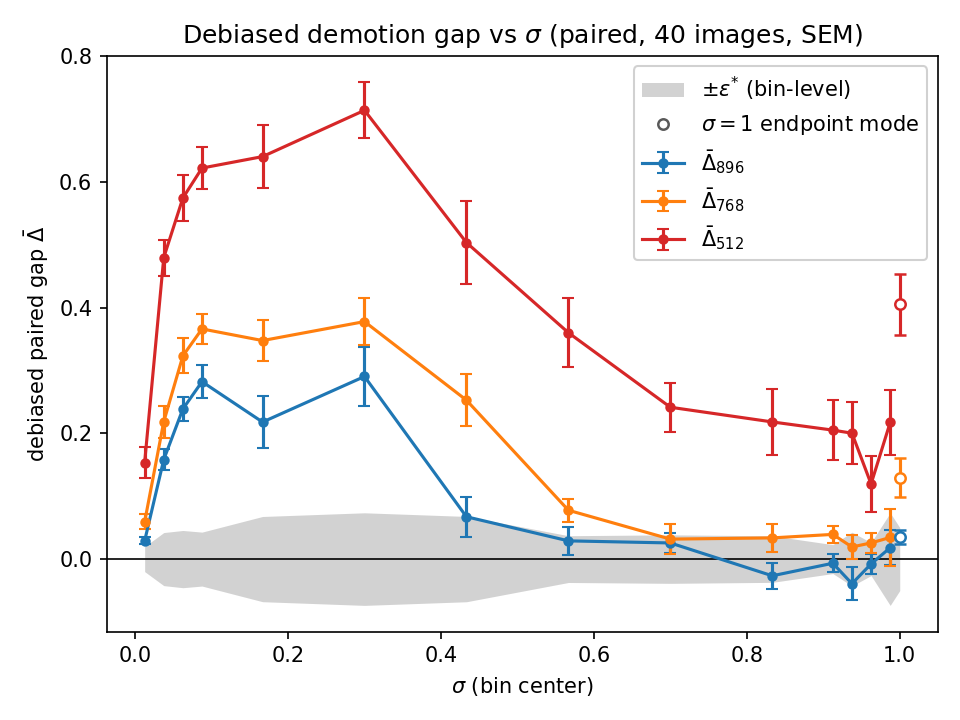}
\phantomcaption
\label{fig:gapcurves}
\end{subfigure}\hfill
\begin{subfigure}[t]{0.515\textwidth}
\includegraphics[width=\linewidth]{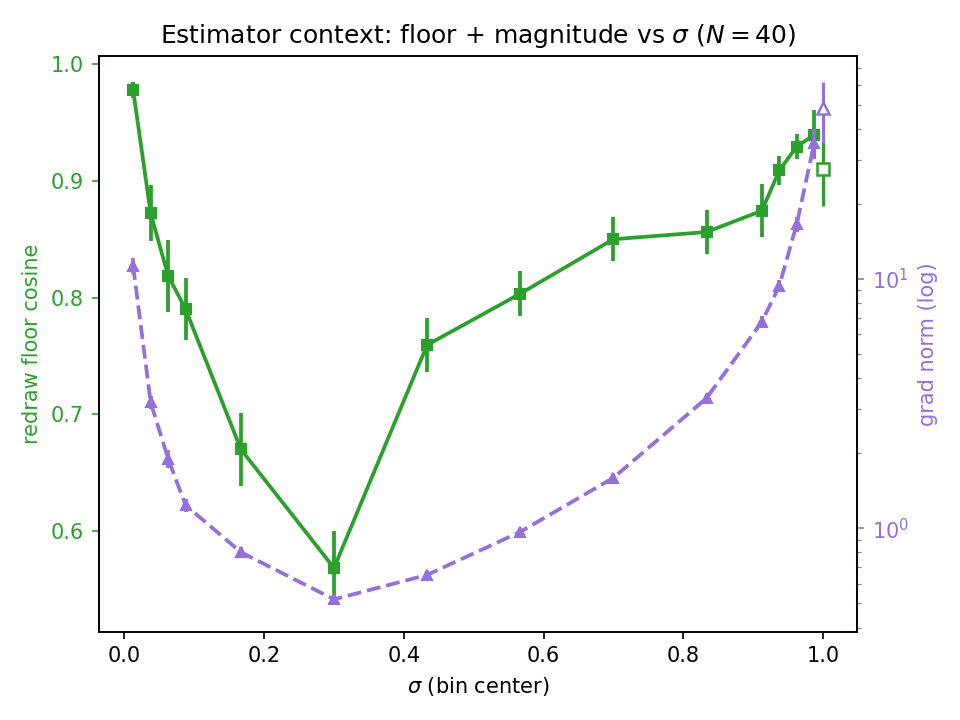}
\phantomcaption
\label{fig:estimator}
\end{subfigure}
\caption{The measurement in detail.
(a)~Debiased demotion gap per $\sigma$-bin, all three routes on one
shared axis, paired against the re-encoding control ($1024$-tier
natives, $N{=}40$; per-bin values and
SEMs in Table~\ref{tab:debiasedmap}; gray band: bin-level
$\pm\epsstar$ (\S\ref{sec:instrument}); open markers: the $\sigma{=}1$
endpoint bin): safety is a per-route boundary. The $896$ route
enters the band at $\sigma \approx 0.5$, $768$ falls only to a
low-excess shelf ($+0.02$--$0.04$ over $0.65 < \sigma < 0.95$, short
of the strict margin), and $512$ never falls inside the band (closest
approach $+0.12 \pm 0.05$ at $\sigma = 0.96$); all three boundaries
stand against the spectral family's prediction of
$\sigma \approx 0.14$--$0.20$.
(b)~The estimator context those curves are read through: the
redraw-floor cosine and the U-shaped gradient norm
$\|\bar g_{\src}(\sigma)\|$.}
\label{fig:verdict}
\end{figure}

Table~\ref{tab:debiasedmap} is the numerical form of
Fig.~\ref{fig:gapcurves}, the paper's verdict map, and
Table~\ref{tab:floor} the debiased floor reads that
\S\ref{sec:ourscored} scores the graph term on. The remaining tables
are the raw-estimator record: retained for continuity and for the reads
defended in \S\ref{sec:instrument} (one-signed bias; paired same-grid
contrasts), not cited as debiased evidence. Table~\ref{tab:phase0} is
the uncorrected counterpart of the verdict map on the original coarse
grid, and its last two rows are the coarse-grid estimator context
(Fig.~\ref{fig:estimator} plots the same two curves on the dense
verdict grid).

\begin{table}[h]
\caption{\textbf{Debiased verdict map} (plotted in
Fig.~\ref{fig:gapcurves}): paired per-image debiased excess
$\bgap$ (SEM) against the re-encoding control, $1024$-tier natives
($N{=}40$, $D{=}12$/bin, deterministic kernels, on a segmented grid
dense below $\sigma = 0.1$ and above $0.9$: $14$ bins in $(0, 1)$ plus
a $\sigma{=}1$ endpoint bin; trimmed per \S\ref{sec:instrument},
retaining $36$--$40$ images per cell). Bin-level $\epsstar$ at this
$(N, D)$ is $0.02$--$0.07$. A reconstruction check (the realized
excess rebuilt from per-arm means) flags the three $\sigma \le 0.06$
bins as partly estimator inflation, so the raw-paired excess is the
primary read there ($896$: $+.037/+.118/+.221$; $768$:
$+.065/+.202/+.275$; $512$: $+.172/+.406/+.497$), up to $0.08$
smaller with the same verdicts.}
\label{tab:debiasedmap}
\centering
\small
\begin{tabular}{cccc}
\toprule
$\sigma$ bin center & $1024{\to}896$ & $1024{\to}768$ & $1024{\to}512$ \\
\midrule
0.013 & $+.029$ {\scriptsize(.006)} & $+.059$ {\scriptsize(.012)} & $+.154$ {\scriptsize(.025)} \\
0.038 & $+.158$ {\scriptsize(.016)} & $+.219$ {\scriptsize(.026)} & $+.480$ {\scriptsize(.029)} \\
0.063 & $+.239$ {\scriptsize(.019)} & $+.323$ {\scriptsize(.028)} & $+.575$ {\scriptsize(.037)} \\
0.088 & $+.282$ {\scriptsize(.026)} & $+.366$ {\scriptsize(.024)} & $+.622$ {\scriptsize(.033)} \\
0.167 & $+.218$ {\scriptsize(.041)} & $+.348$ {\scriptsize(.033)} & $+.640$ {\scriptsize(.050)} \\
0.300 & $+.291$ {\scriptsize(.047)} & $+.378$ {\scriptsize(.038)} & $+.714$ {\scriptsize(.045)} \\
0.433 & $+.067$ {\scriptsize(.032)} & $+.253$ {\scriptsize(.041)} & $+.504$ {\scriptsize(.065)} \\
0.567 & $+.029$ {\scriptsize(.023)} & $+.078$ {\scriptsize(.019)} & $+.360$ {\scriptsize(.055)} \\
0.700 & $+.026$ {\scriptsize(.016)} & $+.032$ {\scriptsize(.023)} & $+.242$ {\scriptsize(.039)} \\
0.833 & $-.027$ {\scriptsize(.020)} & $+.034$ {\scriptsize(.023)} & $+.218$ {\scriptsize(.053)} \\
0.913 & $-.007$ {\scriptsize(.014)} & $+.039$ {\scriptsize(.013)} & $+.205$ {\scriptsize(.047)} \\
0.938 & $-.039$ {\scriptsize(.026)} & $+.019$ {\scriptsize(.019)} & $+.200$ {\scriptsize(.049)} \\
0.963 & $-.008$ {\scriptsize(.015)} & $+.026$ {\scriptsize(.016)} & $+.120$ {\scriptsize(.045)} \\
0.988 & $+.018$ {\scriptsize(.028)} & $+.034$ {\scriptsize(.045)} & $+.217$ {\scriptsize(.052)} \\
\midrule
1.0 (endpoint) & $+.034$ {\scriptsize(.011)} & $+.130$ {\scriptsize(.030)} & $+.406$ {\scriptsize(.049)} \\
\bottomrule
\end{tabular}
\end{table}

\begin{table}[h]
\caption{\textbf{The floor, debiased} (scored in
\S\ref{sec:ourscored}). \textbf{Endpoint}: at $\sigma{=}1$ the
input is exactly $\epsilon$ on every grid, so the input-mediated data
term vanishes by construction. \textbf{x-zero}: the image is removed
from input \emph{and} target; any surviving gap is pure graph-shape
sensitivity. Debiased draw-limit values $\dgap_\infty$ [bootstrap 95\%
CI over images] from nested draw sweeps (endpoint: $N{=}12$,
$D = 4 \ldots 64$; x-zero: $N{=}40$, $D = 4 \ldots 32$). The two
columns agree within CI at every route; our pre-registered
falsification criterion (all endpoint gaps ${\approx}\,0$) is not met;
and no member of the
spectral family can produce a nonzero entry here.}
\label{tab:floor}
\centering
\small
\begin{tabular}{lcc}
\toprule
route & endpoint $\dgap_\infty$ [95\% CI] &
x-zero $\dgap_\infty$ [95\% CI] \\
\midrule
native self-floor (cosine) & $1.005$ $[0.994, 1.016]$ & --- \\
re-encoding control & $-0.003$ $[-0.017, +0.008]$ & --- \\
$1024{\to}896$ & $+0.019$ $[+0.010, +0.030]$ & $+0.034$ $[+0.017, +0.058]$ \\
$1024{\to}768$ & $+0.056$ $[+0.043, +0.071]$ & $+0.074$ $[+0.053, +0.094]$ \\
$1024{\to}512$ & $+0.304$ $[+0.197, +0.424]$ & $+0.283$ $[+0.232, +0.332]$ \\
\bottomrule
\end{tabular}
\end{table}

\begin{table}[h]
\caption{\textbf{Raw estimator} (historical record; superseded by
Table~\ref{tab:debiasedmap}): demotion gap by
$\sigma$-bin, $1024$-tier natives ($N{=}40$,
bin-mean SEM ${\sim}0.02$, re-encoding control $|\gap_\reenc| \le 0.054$
everywhere, split-half reliability $0.73$--$0.83$). Bold = within the
re-encoding band. Raw values understate mid-$\sigma$ gaps (the floor
attenuates the read) and overstate endpoint floors (single-estimate
variance bias); the debiased map corrects both.}
\label{tab:phase0}
\centering
\small
\begin{tabular}{lcccccccc}
\toprule
$\sigma$ bin center & 0.06 & 0.19 & 0.31 & 0.44 & 0.56 & 0.69 & 0.81 & 0.94 \\
\midrule
$\gap_{896}$ (ratio $0.875$) & .110 & .162 & .148 & .137 & \textbf{.048} & \textbf{.030} & \textbf{.030} & .053 \\
$\gap_{768}$ (ratio $0.75$)  & .144 & .216 & .208 & .223 & .164 & .163 & .115 & .063 \\
$\gap_{512}$ (ratio $0.5$)   & .348 & .410 & .355 & .469 & .430 & .391 & .289 & .296 \\
\midrule
$\cos_{\mathrm{floor}}$ & .84 & .65 & .51 & .65 & .70 & .77 & .80 & .83 \\
$\|g\|$ (native) & 2.6 & 0.5 & 0.4 & 0.5 & 0.7 & 1.1 & 2.0 & 9.3 \\
\bottomrule
\end{tabular}
\end{table}

\begin{table}[h]
\caption{Iso-severity probe (\textbf{raw estimator}): gap by $\sigma$-bin
on the $1280$-tier cache
($N{=}24$, $4$ draws/bin) against the corpus $1024{\to}896$ curve. The
ratio-matched routes coincide within ${\sim}1.4$ combined SEM at every bin
despite a $1.6\times$ difference in absolute target size; the same-native
harsher-ratio route separates. $^{*}$$896$'s canonical $16$-draw endpoint
is $-0.009$ raw, $+0.019$ debiased (Table~\ref{tab:floor}).}
\label{tab:isoseverity}
\centering
\small
\begin{tabular}{lccccc}
\toprule
$\sigma$ & 0.125 & 0.375 & 0.625 & 0.875 & 1.0 \\
\midrule
re-encoding control & $+.047$ & $-.005$ & $+.048$ & $+.020$ & $-.012$ \\
$1280{\to}1120$ (ratio $.875$) & .129 & .110 & .054 & $-.012$ & $-.007$ \\
$1280{\to}1024$ (ratio $.80$)  & .170 & .178 & .111 & .002 & .061 \\
$1024{\to}896$ (ratio $.875$)  & .132 & .076 & .077 & .049 & .100$^{*}$ \\
\bottomrule
\end{tabular}
\end{table}

\begin{table}[h]
\caption{Ratio-transfer run (\textbf{raw estimator}): $896$-tier natives
($N{=}40$), pre-registered
bar ``$\gap_{768}$ within the re-encoding band at $\sigma \ge 0.5$'':
\textbf{fails} (residual $0.06$--$0.12$, ${\sim}2\times$ the
$1024{\to}896$ plateau; ${\sim}$half of it is expected estimator bias,
\S\ref{sec:instrument}; the separation is independently confirmed by
the debiased same-grid floors, \S\ref{sec:ourscored}).}
\label{tab:phase1a}
\centering
\small
\begin{tabular}{lcccccccc}
\toprule
$\sigma$ bin & 0.06 & 0.19 & 0.31 & 0.44 & 0.56 & 0.69 & 0.81 & 0.94 \\
\midrule
$896{\to}768$ & .114 & .209 & .168 & .143 & .124 & .056 & .092 & .061 \\
$896{\to}512$ & .318 & .372 & .308 & .340 & .320 & .280 & .329 & .217 \\
\bottomrule
\end{tabular}
\end{table}

\begin{table}[h]
\caption{Route-uniformity (in amplitude) of the mean prediction
residual: cross-grid
excess of $\bar r = \mathbb{E}_\epsilon[\hat v - (\epsilon - x)]$ over
split-half floors (relative $L_2$; same-grid re-encoding control
$\le 0.02$ everywhere). The three main routes coincide within
${\sim}\pm 0.02$ at every $\sigma$ while their gradient floors span $0$ to
$0.3$. This is a norm read; the direction-resolved follow-up
(Appendix~\ref{app:instrument}) shows the underlying mismatch directions
are image- and route-specific.}
\label{tab:residual}
\centering
\small
\begin{tabular}{lccccc}
\toprule
$\sigma$ & 0.125 & 0.375 & 0.625 & 0.875 & 1.0 \\
\midrule
$1280{\to}1024$ & .885 & .825 & .715 & .465 & .360 \\
$1024{\to}896$  & .862 & .808 & .706 & .459 & .378 \\
$896{\to}768$   & .860 & .814 & .715 & .466 & .393 \\
$768{\to}512$   & .897 & .872 & .767 & .508 & .435 \\
\bottomrule
\end{tabular}
\end{table}

\begin{table}[h]
\caption{Depth localization of the floor (\textbf{raw estimator}): mean
per-block gap at
$\sigma{=}1$, early $=$ blocks 0--9, late $=$ 14--27, for the endpoint
(ep) and x-zero (xz) probes. Early blocks carry ${\sim}3\times$ the
late-block gap. The apparent content share (ep $-$ xz) is a late-block
minority effect in raw units; the debiased draw-limit comparison finds
ep $=$ xz within CI at the whole-adapter level
(Table~\ref{tab:floor}). Within a block every module type sits in
$0.22$--$0.31$ at $512/\sigma{=}1$, including the query and key rows of
the fused QKV projection: landing-side uniformity, which is why the
positional share needed an origin-side intervention to find.}
\label{tab:depth}
\centering
\small
\begin{tabular}{lcccc}
\toprule
route & ep early & ep late & xz early & xz late \\
\midrule
$1024{\to}512$ & .357 & .223 & .351 & .125 \\
$1024{\to}768$ & .164 & .085 & .121 & .033 \\
$1024{\to}896$ & .023 & .015 & .044 & .012 \\
\bottomrule
\end{tabular}
\end{table}

\begin{table}[h]
\caption{Positional-interpolation intervention at the $\sigma{=}1$
endpoint ($N{=}40$, $16$ draws; \textbf{raw} arm values; the paired
$\Delta$ column is a same-grid contrast in which the finite-draw bias
cancels to first order): exact phase-geometry alignment erases the
mild-route floor and ${\sim}30\%$ of the harsh-route floor.}
\label{tab:pi}
\centering
\small
\begin{tabular}{lcccc}
\toprule
route & plain (SEM) & PI-aligned (SEM) & paired $\Delta$ (SEM) & improved \\
\midrule
$1024{\to}896$ (control) & $-0.021$ (.041) & $-0.040$ (.040) & --- & --- \\
$1024{\to}768$ & $+0.080$ (.048) & $-0.001$ (.039) & $+0.081$ (.031) & $78\%$ \\
$1024{\to}512$ & $+0.320$ (.058) & $+0.224$ (.056) & $+0.096$ (.039) & $70\%$ \\
\bottomrule
\end{tabular}
\end{table}

\end{document}